\documentclass[sigconf,authordraft, review=false]{acmart}
\usepackage{makecell}
\usepackage{hyperref}
\usepackage{subcaption}

\AtBeginDocument{%
  \providecommand\BibTeX{{%
    \normalfont B\kern-0.5em{\scshape i\kern-0.25em b}\kern-0.8em\TeX}}}

\begin{document}

\title{Reducing Spurious Correlation for Federated Domain Generalization}

\author{Shuran Ma}
\email{shrma@stu.xidian.edu.cn}
\affiliation{%
  \institution{Xidian University}
  \city{Xi'an}
  \state{Shaanxi}
  \country{China}
}

\author{Weiying Xie}
\email{wyxie@xidian.edu.cn}
\affiliation{%
  \institution{Xidian University}
  \city{Xi'an}
  \state{Shaanxi}
  \country{China}
}

\author{Daixun Li}
\email{ldx@stu.xidian.edu.cn}
\affiliation{%
  \institution{Xidian University}
  \city{Xi'an}
  \state{Shaanxi}
  \country{China}
}

\author{Haowei Li}
\email{23011210779@stu.xidian.edu.cn}
\affiliation{%
  \institution{Xidian University}
  \city{Xi'an}
  \state{Shaanxi}
  \country{China}
}

\author{Yunsong Li}
\email{ysli@mail.xidian.edu.cn}
\affiliation{%
  \institution{Xidian University}
  \city{Xi'an}
  \state{Shaanxi}
  \country{China}
}



\begin{abstract}
The rapid development of multimedia has provided a large amount of data with different distributions for visual tasks, forming different domains. Federated Learning ($FL$) can efficiently use this diverse data distributed on different client media in a decentralized manner through model sharing. However, in open-world scenarios, there is a challenge: global models may struggle to predict well on entirely new domain data captured by certain media, which were not encountered during training.
Existing methods still rely on strong statistical correlations between samples and labels to address this issue, which can be misleading, as some features may establish spurious short-cut correlations with the predictions. To comprehensively address this challenge, we introduce $FedCD$ (\textbf{C}ross-\textbf{D}omain Invariant \textbf{Fed}erated Learning), an overall optimization framework at both the local and global levels. We introduce the \textbf{S}purious \textbf{C}orrelation \textbf{I}ntervener ($SCI$), which employs invariance theory to locally generate interventers for features in a self-supervised manner to reduce the model's susceptibility to spurious correlated features. Our approach requires no sharing of data or features, only the gradients related to the model. Additionally, we develop the simple yet effective \textbf{R}isk \textbf{E}xtrapolation \textbf{A}ggregation strategy ($REA$), determining aggregation coefficients through mathematical optimization to facilitate global causal invariant predictions. Extensive experiments and ablation studies highlight the effectiveness of our approach. In both classification and object detection generalization tasks, our method outperforms the baselines by an average of at least 1.45\% in $Acc$, 4.8\% and 1.27\% in $mAP_{50}$. The code is released at:xxx.
\end{abstract}






\maketitle
\section{Introduction}
The swift evolution of multimedia technology has provided us with a rich and diverse data resource, greatly advancing the development of visual tasks. The diversity of multimedia data is reflected not only in the wide range of artistic forms, such as visual arts, photography, and animation but also in the different environments considering the autonomous driving scenarios, such as scenes and weather conditions. These diverse data samples are valuable materials that have significantly enhanced the ability of neural networks to understand and process visual information.
Faced with strict requirements for privacy protection, traditional centralized learning methods are limited by the risk of privacy leakage, making it difficult to utilize multimedia data. Federated Learning ($FL$), as an emerging machine learning framework, provides an effective solution. $FL$ allows multiple devices or medias to train models locally and improve the global model collaboratively by sharing model updates or gradients instead of raw data.
The rapid rise of new media continuously introduces new domain data, which poses requirements for domain generalization, referring to the ability of a model to maintain good performance when faced with new domain data distributions that differ from the training set. Although federated learning facilitates the joint training of different media data, it is not sufficient to address this issue, as it still relies on simple collaborative training of client data and may struggle to predict on unseen domains. 

\begin{figure}
	\centering
	\includegraphics[width=0.5\textwidth]{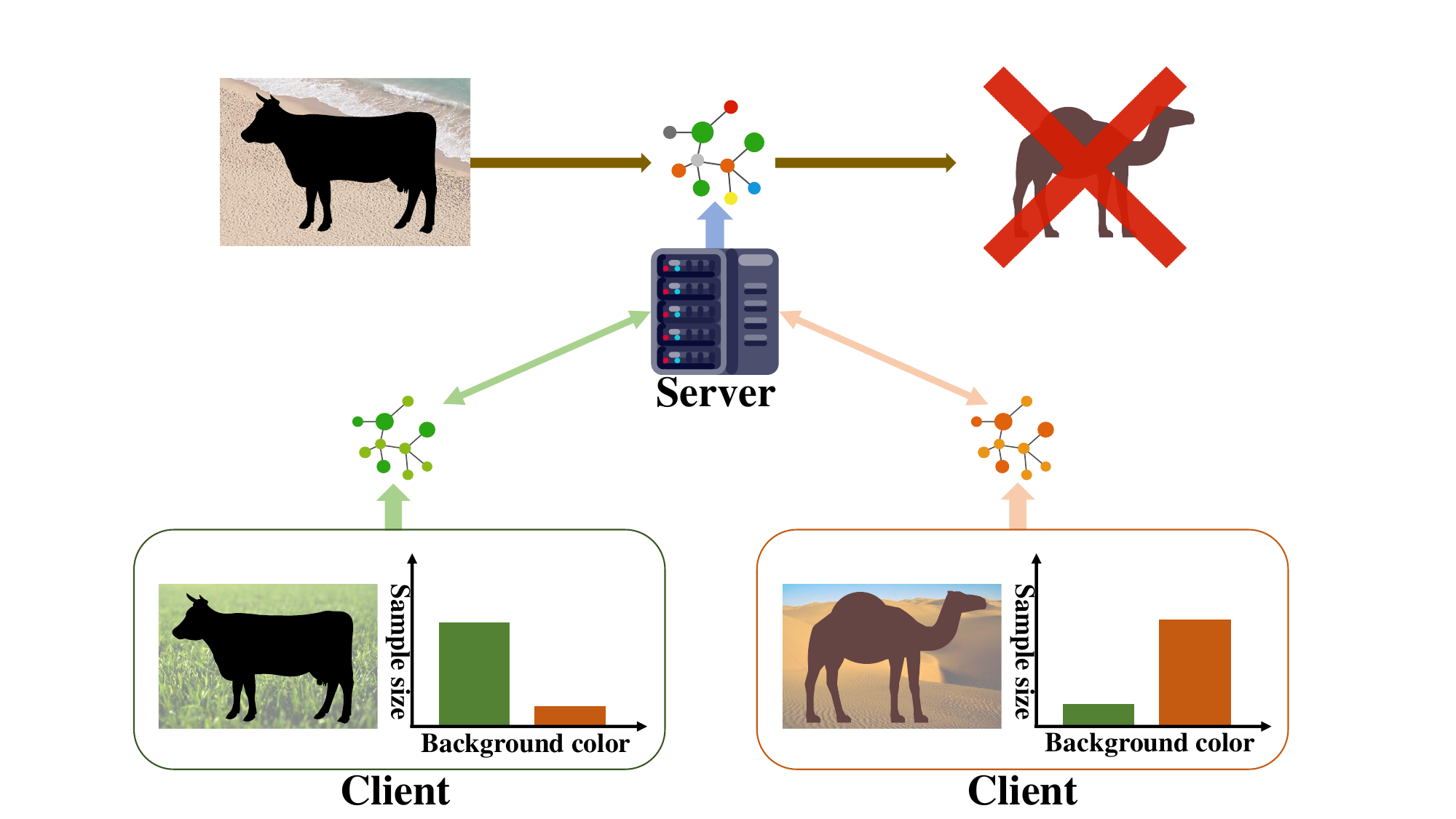}
	\caption{The "Cow-On-The-Beach" problem in $FL$. Local models on the client side strongly correlate cows with green grassland and camels with beige deserts within the samples. Thus, replacing the background of the cow with a beige beach may lead to incorrect prediction on global model.}
	\label{fig: 1cowonthebeach}
\end{figure}

There have been some early works aiming to address the aforementioned issue. Representation learning was utilized in~\cite{pmlr-v189-sun23a,NEURIPS2022_fd946a6c,9710573} to tackle this challenge by considering various aspects such as conditional mutual information and the distribution of features across different clients. Liu \textit{et al.} employed meta-learning which is considered programming-unfriendly in~\shortcite{liu2021feddg}. Additionally, this work shared amplitude spectrum which is data-related, potentially contradicting the principles of federated learning. Qu \textit{et al.}\shortcite{qu2022generalized} improved domain generalization by providing a flatter, more generalized loss function. However, these methods still utilized all features unbiasedly, which means they used all features without intervention and allowed the neural network to establish connections between any features and predictions based solely on statistical correlations, without considering that some of these correlations might be spurious shortcuts.
However, we think they are insufficient because only a subset of the target's features in the sample ultimately leads to predictions, excluding features related to the scene, style or other aspects of the target, such as the color of the car. Therefore, we should not consider all of them when making predictions, or at least not unbiasedly, which can provide a more thorough understanding and stable performance.

To achieve a more comprehensive improvement, we will delve deeper into the spurious correlations between data features and predictions. As shown in Figure \ref{fig: 1cowonthebeach}, inspired by the "cow-on-the-beach" example presented in~\cite{arjovsky2019invariant}, neural networks rely on simple statistical relationships, e.g., cow and green grass to make predictions, which establishes a strong correlation between a green background and a cow, leading to wrong predictions when the background changes to a beige beach. We believe that the model's capture of spurious "short-cut" features hinders generalization. Inspired by some non-causal and invariance theories~\cite{xu2023multi, koyama2021invariance} that reduce spurious correlations, we locally design a \textbf{S}purious \textbf{C}orrelation \textbf{I}ntervener ($SCI$) to mitigate the impact of spurious features on model training. Additionally, leveraging causal invariance theory~\cite{arjovsky2019invariant,krueger2021out}, we propose the \textbf{R}isk \textbf{E}xtrapolation \textbf{A}ggregation strategy ($REA$) from a causal inference perspective.

Within the $SCI$, we reference the principle of invariance to design a self-supervised feature intervener, aiming to reduce the impact of spurious correlated features. Moreover, the training of $SCI$ only requires the sharing of model-related gradients, without violating the principles of federated learning, and is even more lightweight than sharing data-related information. In light of insights from~\cite{arjovsky2019invariant,krueger2021out}, which affirm the relationship between training risk and the causal invariance mechanism of domain generalization, we propose the simple yet effective $REA$ strategy during aggregation. This strategy transforms the fixed aggregation coefficients into a set of solutions for the risk extrapolation optimization problem, further enhancing global invariant causal predictions.

Our main contributions are highlighted as follows:
\begin{itemize}
	\item We aim to address the federated domain generalization problem by considering spurious correlations. We locally and self-supervisedly train feature interveners for each client, while also considering the strong prior dependence on causal invariant components of images, and the constraints thereby imposed. Additionally, we provide a detailed theoretical derivation of the training objective. 
\end{itemize}

\begin{itemize}
	\item We introduce $REA$, a novel aggregation strategy that recalculates aggregation coefficients through a simple yet effective risk extrapolation optimization problem which does not require the introduction of new network structures for solving, making it efficient and easy to implement.
\end{itemize}

\begin{itemize}
	\item Our method only requires the sharing of an additional model-related gradient, without any information from the data or features themselves. Comprehensive comparative experiments and ablation results demonstrate the effectiveness of our method.
\end{itemize}

\section{RELATED WORK}
\subsection{Federated Learning}
Federated Learning is a paradigm for training models using decentralized client data while preserving privacy. Existing mature explorations have primarily focused on scenarios with heterogeneous client data, such as class imbalance in~\cite{shen2021agnostic,ye2023fedfm,yoon2021federated,li2022federated}. Some efforts have concentrated on modifying local training in~\cite{li2020federated,li2021model,mu2023fedproc}, with~\cite{li2020federated} adding L2 regularization between local and global models, \cite{li2021model,mu2023fedproc} performing contrastive learning locally. Other works~\cite{NEURIPS2020_564127c0,dong2023fed,uddin2023arfl} have rebuilt the aggregation strategy. All these approaches share a common goal, which is to tackle the challenge of training drift among clients caused by data heterogeneity. It's essential to note that this problem differs from domain generalization. The distribution of the test set in the former case remains a subset of that in the training set, which indicates it is already encountered by the model. In contrast, domain generalization extends to the test set from unseen domains (e.g., weather, scenes), where the distributions are unknown. While Yuan \textit{et al.}~\shortcite{yuan2021we} emphasized the importance of both types of research, our primary focus lies on domain generalization perspective. Because in the realm of multimedia, domain generalization presents a critical challenge owing to the high diversity and complexity of multimedia data. In open-world scenarios, the global model frequently encounters previously unseen data domains due to various factors such as collection environments, equipment disparities, and processing methodologies. Consequently, the model must possess robust generalization capabilities to accurately predict outcomes in the presence of these unfamiliar domains.
\subsection{Domain Generalization}
Domain generalization aims to enable models to make accurate predictions on domains that were not seen during training. Approaches in~\cite{li2018domain,li2018deep,muandet2013domain,shao2019multi,shi2021gradient,wang2020learning} minimized domain gaps or learned domain-invariant features to achieve domain generalization. Meta-learning methods in~\cite{NEURIPS2018_647bba34,NEURIPS2019_2974788b,li2018learning} enhanced generalization capabilities by training models to "learn how to learn", but they were not programming-friendly. However, most of the above centralized methods require domain labels or direct access to multi-source domain data, which is constrained in the context of federated learning.
\subsection{Federated Domain Generalization}
Federated learning domain generalization is an emerging challenge that requires the global model to perform well on an unseen domain. Liu \textit{et al.}~\shortcite{liu2021feddg} addressed it by sharing amplitude spectra-rich domain-related information, which is data-related and may lead to privacy leakage. Xu \textit{et al.}~\shortcite{xu2022closing} created a novel model selector to determine the closest model/data distribution for any test data. Zhang \textit{et al.}~\shortcite{zhang2023federated} focused on aggregation, and improved the global model's generalization by reducing generalization differences between global and local models and increasing flatness in each domain. These methods still consider all features for prediction unbiasedly, which may lead the model to establish spurious short-cut correlations between data features and predictions. Consequently, when the model encounters unseen domain data, it may make incorrect predictions due to these confounders, even though the nature of the target remains unchanged. Therefore, we propose to intervene at the feature level during the federated training process by self-supervisedly training feature interveners to mitigate the adverse effects caused by spurious components. In the aggregation process, we further optimize the global model's cross-domain invariant prediction ability based on the theory of causal invariance.

\section{PROPOSED METHOD}

\subsection{Problem Setting}
\textbf{Motivation. }The generalization ability of $FL$ is crucial for addressing open-world multimedia problems and advancing general artificial intelligence. Invariant Risk Minimization ($IRM$) theory suggests in~\cite{arjovsky2019invariant} that models often capture spurious correlations in data, thereby impairing their ability to generalize. In the context of structural causal models, this issue arises from spurious correlations induced by data selection bias, where the true influences should come from causal invariant features.

The settings of $FL$ have pros and cons for addressing this issue. Firstly, its diverse client-side heterogeneous data distributions can provide the global model with insights into a wider range of domains. However, its privacy-preserving nature means that these domain features related to the data are not easily shared like in a centralized manner.

Additionally, many centralized methods for removing spurious correlations are based on \textbf{strong prior knowledge} and have \textbf{constraints}. For example, methods that remove style information confounder~\cite{jin2021style} and background information confounder~\cite{wang2021proactive, shao2021improving} are based on the strong prior assumption that these types of information should not be causally related to the prediction.
However, their constraints lie in the fact that their specificity to the objectives based on the above priors, thus can only remove the influence of one type of spurious correlated features at a time. It was also pointed out in~\cite{xu2023multi} that the inadequacy of prior-based methods and found that causal features $s_{cau}$ should actually be included in domain-invariant ones $s_{com}$, while domain-private features $s_{pri}$, which is variant, are contained in non-causal ones $s_{non}$, so there is:
\[
\left\{
\begin{array}{l}
	\text{$s_{cau} \subset s_{com}$}, \\
	\text{$s_{non} \supset s_{pri}$}.
\end{array}
\right.
\]

Therefore, when seeking to address the federated domain generalization problem by tackling spurious correlations, we need to avoid strong prior dependence on causal invariant components of images and the resulting constraints as much as possible. Thus, we directly apply the theory of invariance on the feature space to obtain a more general approach.

\textbf{Formulation. }
In the federated learning framework, we consider that each client's data corresponds to a specific environment $\epsilon$. All clients together form the training environment set for federated learning, denoted as $\mathcal{E}_{tr} = \{\epsilon_1, ...\epsilon_e,...\epsilon_E\}$, where E corresponds to the total number of clients. The data distribution of the $e$-th client is represented by $P(\mathcal{X}_e, \mathcal{Y}_e | \epsilon_e)$, where $\mathcal{X}_e$ and $\mathcal{Y}_e$ respectively denote the raw data and labels of client $e$ with environment $\epsilon_e$. Each client conducts local training with the objective of $\mathop{\arg\min}\limits_{\theta} \mathcal L_e(\mathcal X_e, \theta_e^r)$ in communication round $r(1 \leq r \leq R)$. Subsequently, clients upload their local models $\theta_e^r$ to the server, where aggregation $\theta^r = \sum_{e=1}^{E}\frac{N_e}{N}\theta_e^r$ is performed to obtain the global model $\theta^r$ in round $r$, where $N_e$ denotes the number of samples on client $e$ that participate in the training, and $N$ denotes total number of samples involved in training across all clients. Therefore, the global objective of federated learning can be expressed as:
\begin{align}
	\mathop{\arg\min}\limits_{\theta} \sum_{e=1}^{E}\frac{N_e}{N}\mathcal L_e(\mathcal X_e,\theta). \label{fedavg agg}
\end{align}%

For domain generalization, Eq.~\ref{fedavg agg} will undergo some changes. The global model needs to make accurate predictions in the environment $\epsilon_{e'} \in \mathcal{E}' = \{\epsilon_{1'}, \ldots, \epsilon_{e'}, \ldots, \epsilon_{E'}\}$ of unseen domains. This requires the model to generalize well across diverse and previously unseen environments, with the goal of:
\begin{align}
	\mathop{\arg\min}\limits_{\theta} \mathop{\max}\limits_{\epsilon_{e'} \in \mathcal{E}'}\mathcal{L}(f(\mathcal{X}_{e'};\theta), \mathcal{Y}_{e'} | \epsilon_{e'}). \label{feddg}
\end{align}%
Where $f(\mathcal{X}_{e'};\theta)$ represents the prediction obtained using the global model $\theta$ on data $\mathcal{X}_{e'}$, $\mathcal{L}$ quantifies the error between this prediction and the true label $\mathcal{Y}_{e'}$.

\subsection{Invariant Optimization Theory}
We propose an efficient module, namely $SCI$, based on the principles of invariance to reduce spurious correlations at the client level, and provide a theoretical derivation for this purpose.

\textbf{Optimization objective. }It was suggested in~\cite{koyama2021invariance} that causality-based out-of-distribution ($OOD$) studies, which include domain generalization issues, imply the existence of a feature subset $F^{S} = \{F^{s}; s \in S\}$ such that the predictions generated using these features exhibit invariance across domains, i.e., they are independent of the environment $\epsilon$, $P(Y|F^S, \epsilon_i) = P(Y|F^S, \epsilon_{j \neq i}) = P(Y|F^S)$.

In our approach, we aim to design interveners for features to reduce spurious correlations by weakening the feature components to varying degrees, that is, $F^S := M_{\delta}(X)$, where $M_{\delta}(\cdot)$ is $\delta$ parameterized and acts as a spurious feature intervener. So our optimization objective is to minimize the KL divergence:
\begin{align}
 d_{KL}[P(Y|M_{\delta_{e}}(X), \epsilon_{e}) || P(Y|M_{\delta_{e}}(X))], \forall \epsilon_{e} \in \mathcal{E}_{tr}, 
\end{align}%
and can be further written as:
\begin{align}
	\mathop{\arg\min}\limits_{\delta_e} \mathbb{E}_{\mathcal{E}_{tr}}[d_{KL}(P(Y|M_{\delta_{e}}(X), \mathcal{E}_{tr}) || P(Y|M_{\delta_{e}}(X)))], \label{mask_obj}
\end{align}%
which is optimized through the learnable parameter $\delta_{e}$ for every client with domain environment $\epsilon_{e} \in \mathcal{E}_{tr}$.

\textbf{Invariance-based simplification. }
We still cannot obtain the relationship between the available intervener optimization objective and the learnable parameter $\delta$ to generate it only through Eq.~\ref{mask_obj}. Fortunately,~\cite{koyama2021invariance} further manifests this connection in a centialized setting as shown in Eq.~\ref{to var grad}.
\begin{equation}\label{to var grad}
	\begin{aligned}
		&\mathbb{E}_{\mathcal{E}_{tr}}[d_{KL}(P(Y|M_{\delta}(X), \mathcal{E}_{tr}) || P(Y|M_{\delta}(X)))],\\
		=~& \mathbb{E}_{\mathcal{E}_{tr}}[log~P(Y|M_{\delta}(X), \mathcal{E}_{tr}) - P(Y|M_{\delta}(X))], \\
		=~& \alpha(\mathbb{E}_{\mathcal{E}_{tr}}[\nabla_\delta\mathcal{L}_{\mathcal{E}_{tr}}(\delta)^T\nabla_\delta\mathcal{L}_{\mathcal{E}_{tr}}(\delta)] \\ &- \mathbb{E}_{\mathcal{E}_{tr}}[\nabla_\delta\mathcal{L}_{\mathcal{E}_{tr}}(\delta)]^T\mathbb{E}_{\mathcal{E}_{tr}}[\nabla_\delta\mathcal{L}_{\mathcal{E}_{tr}}(\delta)]) + O(\alpha^2),\\
		=~&\alpha~trace(Var_{\mathcal{E}_{tr}}(\nabla_{\delta}\mathcal{L}_{\mathcal{E}_{tr}}(\delta))) + O(\alpha^2),
	\end{aligned}%
\end{equation}
where $\mathcal{L}_{\mathcal{E}_{tr}}(\delta) = \mathbb{E}_{(X, Y)\in\mathcal{E}_{tr}}[l(Q(Y|X;\delta), Y)|\mathcal{E}_{tr}]$ can be considered as the average predicted loss after using $\delta$ to generate intervener for all data $(X, Y) \in \mathcal{E}_{tr}$ of the whole training set, $Q(\cdot)$ is the corresponding black-box system and only appears as an intermediate variable in the derivation. 
$\alpha$ can be considered as the learning rate for $\delta$, and we have $\delta' = \delta - \alpha\nabla_\delta\mathcal{L}_{\mathcal{E}{tr}}(\delta)$.
It is important to note that our setup differs from the centralized methods described above in two key aspects:

\hypertarget{aspect 1}{1. }In the centralized setup, the environment set $\mathcal{E}_{tr}$ is considered as a whole. However, in the federated setting, each $\epsilon_e$ corresponds to an individual client $e$. Therefore, we need to emphasize again that $\epsilon_e \in \mathcal{E}_{tr} =\{\epsilon_1, ..., \epsilon_e, ..., \epsilon_E\}$, meaning that we approach this problem from a client-wise perspective.

\textbf{2.} The centralized method described above can train a single $\delta$ without distinguishing the heterogeneity of each $\epsilon_e$ in $\mathcal{E}_{tr}$, as they are considered as a whole. 
However, this is not applicable to our setup as described in aspect~\hyperlink{aspect 1}{1}, because $\mathcal{E}_{tr}$ cannot be accessed directly and in its entirety by any media. Thus, we generate a personalized $M_{\delta_e}(\cdot)$ with learnable parameter $\delta_e$ for each client $e$ using Eq~\ref{mask_obj}. Additionally, all averaging operations $\mathbb{E}_{\mathcal{E}_{tr}}$ are replaced with client-wise averaging $\mathbb{E}_c$. Therefore, the overall train-set average loss is modified to be intra-client, as shown in Eq.~\ref{env wise loss}.
\begin{align}\label{env wise loss}
	\mathcal{L}_{\epsilon_e}(\delta_e) = \mathbb{E}_{(X_e, Y_e)\in\epsilon_e}[l(Q(Y_e|X_e;\delta_e), Y_e)|\epsilon_{e}],~\epsilon_e \in \mathcal{E}_{tr}.
\end{align}%
For simplicity, the gradient of the $\mathcal{L}_{\epsilon_e}(\delta_e)$ is denoted as:
\begin{align}\label{simple grad}
	\nabla_{\delta_e}\mathcal{L}_{\epsilon_e}(\delta_e)~:=~\nabla_{e},~e \in [1, ..., e, ..., E],
\end{align}%
and each $\nabla_e$ is considered separately.

Therefore, we take the intermediate term from Eq.~\ref{to var grad} to create a new optimization objective:
\begin{equation}\label{new opt}
	\begin{aligned}
		min ~& \underbrace{\mathbb{E}_c[\nabla_{\delta_e}\mathcal{L}_{\epsilon_e}(\delta_e)^T\nabla_{\delta_e}\mathcal{L}_{\epsilon_e}(\delta_e)]}_{term~1},\\
		&-\underbrace{\mathbb{E}_c[\nabla_{\delta_e}\mathcal{L}_{\epsilon_e}(\delta_e)]^T\mathbb{E}_c[\nabla_{\delta_e}\mathcal{L}_{\epsilon_e}(\delta_e)]}_{term~2},
	\end{aligned}%
\end{equation}
and derive these two terms separately.
\begin{figure}
	\centering
	\includegraphics[width=0.47\textwidth]{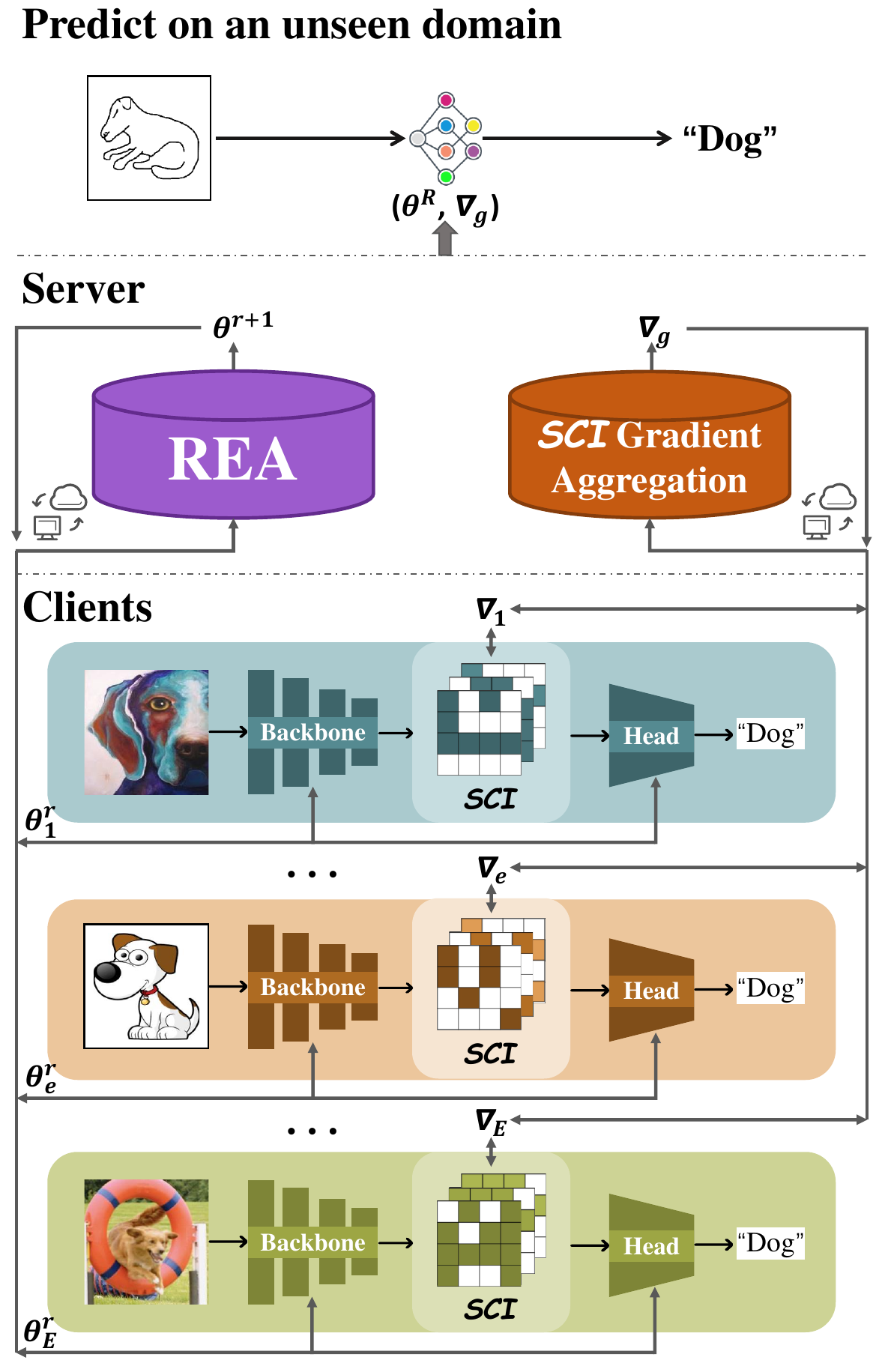}
	\caption{A brief overview of $FedCD$, consisting of two components $SCI$ and $REA$. In $SCI$, each client $e$ compute a gradient alignment penalty using Eq.~\ref{img loss} and upload the local $SCI$ gradient tensors $\nabla_e$ in each communication round. They are then aggregated on the server to obtain $\nabla_g$, which is returned to the clients for the next round. $REA$ determines aggregation coefficients through mathematical optimization based on Empirical Risk Minimization ($ERM$). Instead of introducing new neural network solutions for optimization, $REA$ employs sequential least squares.}
	\label{fig: 5FrameWork.pdf}
\end{figure}
For the first term, we can intuitively obtain:
\begin{equation}\label{term1 simplify}
	\begin{aligned}
		&\mathbb{E}_c[\nabla_{\delta_e}\mathcal{L}_{\epsilon_e}(\delta_e)^T\nabla_{\delta_e}\mathcal{L}_{\epsilon_e}(\delta_e)], \\
		=~&\mathbb{E}_c[\nabla_e^T\nabla_e],\\
		=~&\sum_{e = 1}^{E}p_e\nabla_e^2,
	\end{aligned}%
\end{equation}
here we adopt the simplified form of ~\ref{simple grad}, and concisely write $\nabla_e^T\nabla_e$ as $\nabla_e^2$. $p_e$ represents the probability of each client, serving as the aggregation coefficient of $FedAvg$, i.e., $p_e = \frac{N_e}{N}$.

For the second term, we have:
\begin{equation}\label{term2 simplify}
	\begin{aligned}
		&\mathbb{E}_c[\nabla_{\delta_e}\mathcal{L}_{\epsilon_e}(\delta_e)]^T\mathbb{E}_c[\nabla_{\delta_e}\mathcal{L}_{\epsilon_e}(\delta_e)],\\
		=&~\mathbb{E}_c[\nabla_e | 1 \leq e \leq E]^T\mathbb{E}_c[\nabla_e | 1 \leq e \leq E],\\
		=&~(\sum_{e = 1}^{E}p_e\nabla_e)^T(\sum_{e = 1}^Ep_e\nabla_e),\\
		=&~\nabla_g^2.
	\end{aligned}%
\end{equation}
It can be observed that the penultimate term in Eq.~\ref{term2 simplify} bears similarities to federated aggregation. Thus we define the global $SCI$ gradient $\nabla_g$ as $\nabla_g := \sum_{e = 1}^{E}p_e\nabla_e$.

Therefore, the global objective of $SCI$ is updated to Eq.~\ref{simplify global obj},
\begin{equation}\label{simplify global obj}
	\begin{aligned}
		min ~&|\sum_{e = 1}^{E}p_e\nabla_e^2 - \nabla_g^2|.
	\end{aligned}%
\end{equation}
The absolute value is taken here because the original objective using $d_{KL}$ is always greater than zero.

Since $p_e$ is a fixed constant, we can consider that each client has the optimization objective:
\begin{equation}\label{client obj}
	\begin{aligned}
		min ~&|\nabla_e^2 - \nabla_g^2|.
	\end{aligned}%
\end{equation}
This requires two tensor multiplication operations. To streamline computations, we aim to reduce it to a single tensor multiplication by analyzing the specific form of $\nabla_g$, and there is:
\begin{equation}\label{client obj}
	\begin{aligned}
		&|\nabla_e^2 - \nabla_g^2| = |\nabla_e-\nabla_g|\cdot|\nabla_e+\nabla_g|,\\
		=&~|\nabla_e-\nabla_g|\cdot|\sum_{i = 1, i \neq e}^Ep_e\nabla_e + (1+p_e)\nabla_e|,\\
		=&~|\nabla_e-\nabla_g|\cdot|\sum_{i = 1, i \neq e}^Ep_e\nabla_e + \beta\nabla_e|,\\
		\rightarrow&|\nabla_e-\nabla_g|\cdot|\sum_{i = 1, i \neq e}^Ep_e\nabla_e + (p_e-1)\nabla_e|_{\beta = (p_e - 1)},\\
		=&~|\nabla_e-\nabla_g|\cdot|\nabla_e - \sum_{i = 1}^Ep_e\nabla_e|,\\
		=&~|\nabla_e-\nabla_g|^2.
	\end{aligned}%
\end{equation}

Because $(1 + p_e)$ is a constant, we can treat it as a coefficient $\beta$. Then, we set $\beta = (p_e-1)$ and derive the final form based on the properties of absolute value. We continue to use the simplified notation $|\nabla_e-\nabla_g|^2 := |\nabla_e-\nabla_g|^T|\nabla_e-\nabla_g|$ here.

Finally, the local training loss for $SCI$ is in the form of the L2 norm, requiring only one tensor multiplication:

\begin{align}
	L_{SCI} = ||\nabla_e-\nabla_g||^2\label{img loss}.
\end{align}%
Note that here $\nabla_e$ represents the gradient of the loss function with respect to $M_{\delta_e}(\cdot)$, which is the gradient generated by the optimizer for learning the parameters $\delta_e$, and is different from that of the model parameters in the neural network.

In practical applications, $M_{\delta_e}(\cdot)$ can take on various forms. Here, for computational efficiency, we consider a straightforward approach of feature masking, defining $M_{\delta_e}(X) := M_e \odot X$, where $\odot$ represents element-wise multiplication. 
Additionally, previous methods based on sharing often require each client to upload multiple components related to the data, forming a bank on the server for exchange, such as amplitude spectra ~\cite{liu2021feddg, lv2022causality}, image features~\cite{Chen_2023_WACV}, leading to communication and memory costs. In our approach, regardless of the data distribution, each client only needs to share a single gradient information, making it more lightweight.
\subsection{SCI Implement}
For each client $e$, the final training loss is:
\begin{align}
	L_e = L + \lambda L_{SCI}, \label{client loss}
\end{align}%
where $\lambda$ is a hyperparameter and $L$ depends on the form of downstream tasks, such as classification loss or object detection loss. As shown in Figure~\ref{fig: 5FrameWork.pdf}, each client $e$ has one or more intervener optimizers for generating invariant masks. In each communication round, clients upload the $SCI$ gradients $\nabla_e$, which are aggregated on the server using the $FedAvg$ strategy to obtain $\nabla_g$, which is then distributed to all clients for the next round of computation.

Additionally, $M_{\delta_e}(\cdot)$ is self-supervised generated, and there is no need for prior knowledge about causal invariance information. So it is not specifically designed for a particular type of confounder, but rather for the entire feature space, and thus can remove more types of confounders, i.e., spurious correlations.

\subsection{REA Strategy}

In the aforementioned efforts, we primarily focused on tasks conducted locally on the clients to enhance the domain generalization performance. Next, aiming to improve the global model's causal-invariant predictions, we introduce a novel aggregation strategy $REA$ that determines the aggregation coefficients for the global model through solving an optimization problem related to risk extrapolation, which does not require new neural networks on the server, and is instead solved through sequential least squares.

The standard approach to solve the issue through risk extrapolation is \textbf{E}mpirical \textbf{R}isk \textbf{M}inimization ($ERM$). $ERM$ involves minimizing the average training risks across all domains as shown in Eq. \ref{ERM}, where $\ell$ is the loss function, often assumed to be fixed across different domains, and $\theta$ denotes the model parameters of the neural network. $\mathcal{R}{e_q}(\theta)$ represents the risk of environment $e_q$, and $D{eq}(x, y)$ represents the corresponding data,
\begin{align}
	\mathcal{R}_{ERM}(\theta)&=\sum\limits_{q = 1}^{Q}|D_{e_q}|\mathbb{E}_{(x,y) \sim D_{e_q}}\mathcal{\ell}(f_\theta(x),y)\label{ERM}.
\end{align}

In summary, this approach has two objectives:

1. Reduce training risk.

2. Enhance the similarity of training risks among domains.

Within the framework of federated learning, the first objective can be achieved through local learning. During the aggregation process, our focus is primarily on the second point. Krueger \textit{et al.}~\shortcite{krueger2021out} suggested that optimizing the model using cross-domain risk variance can lead to a flatter "risk plane". Additionally, it provides a smoother gradient vector field during training, meaning the vectors of its loss function can curve more smoothly toward the origin. Compared to other methods such as \textbf{I}nvariant \textbf{R}isk \textbf{M}inimization ($IRM$) in~\cite{arjovsky2019invariant}, this smoother optimization landscape offers robust improvements. Therefore, we design a novel aggregation strategy.

Following the same symbol definitions, $\mathcal{R}{e}(\theta_e^{r})$ denotes the training risk of each client, where we use $L_e$ in Eq.~\ref{client loss}. Assuming aggregation coefficients $w = \{w_1,...,w_E\}$, we can straightforwardly consider the client risks mapped globally as $\mathcal{R}{g_e}(\theta_e^{r}) = w_e\mathcal{R}_{e}(\theta_e^{r})$, similarly to how the global loss is considered in Eq. \ref{fedavg agg}.

Hence, the aggregation coefficients correspond to the solution of the following risk extrapolation optimization problem as:

\begin{align}
	\min_{w}&~Var(\mathcal{R}_{g_1}(\theta_1^{r}), \mathcal{R}_{g_2}(\theta_2^{r}), ..., \mathcal{R}_{g_E}(\theta_E^{r}))\label{Problem},\\
	\text {s.t.}&~\mathcal{R}_{g_e}(\theta_e^{r}) = w_e\mathcal{R}_{e}(\theta_e^{r})\tag{\ref{Problem}{a}} \label{Problema},\\
	&~\sum_{e=1}^{E}w_e = 1\tag{\ref{Problem}{b}} \label{Problemb},\\
	&~w_e \textgreater 0\tag{\ref{Problem}{c}} \label{Problemc},\\
	&~1 \leq e \leq E\tag{\ref{Problem}{d}} \label{Problemd}.
\end{align}

To account for the impact of sample sizes on the global model and combine $w$ with the aggregation coefficients $p_e$ of $FedAvg$, we normalize them to obtain the final aggregation coefficients $\mathcal{C} = \{c_1, ..., c_e, ..., c_E\}$ as follows:

\begin{align}
	c_e&=\frac{exp(\eta{w_e}+p_e)}{\sum\limits_{k = 1}^{K}{exp(\eta{w_e}+p_e)}}\label{final agg coefficient},
\end{align}
where $\eta$ is hyperparameter, denoting the contribution of $ERM$.
\section{EXPERIMENTS}
\subsection{Experimental Setup}
\textbf{Datasets. }
We conducted simulations on both classification tasks with $ResNet18$~\cite{He_2016_CVPR} network and object detection tasks with $YOLOv5$ network using different setups. For the classification tasks, we mainly simulated style generalization on the \textit{PACS}~\cite{li2017deeper} dataset. As for the object detection tasks, we used three autonomous driving datasets, namely \textit{BDD100K}~\cite{yu2020bdd100k}, \textit{Cityscapes}~\cite{cordts2016cityscapes}, and \textit{Mapillary}~\cite{antequera2020mapillary}, to simulate dataset generalization, which exhibits larger domain shifts compared to other configurations. We also simulated scene generalization, especially on the $BDD100K$ dataset.

\textbf{Implemented details. }Our proposed method is implemented in PyTorch, and experiments are conducted with different values for $\lambda = \{0.5, 0.7, 0.8, 0.9, 1.0\}$, and $\eta = \{0.3, 0.5, 0.8\}$, with results provided for the optimal parameter configurations. 

Each mask $M_{\delta_e}(\cdot)$ is initialized to an all-1 tensor, or all-pass, to avoid removing too many features at the beginning of training, which could pre
vent the model from learning correctly.

For the object detection task, we run the experiments on a workstation with NVIDIA A100 GPUs. We set the batch size to 4, conduct a total of 200 communication rounds, and perform 1 local training epoch in each round. The learning rate is set to $1.6\times10^{-4}$.

For the classification task, we run the experiments on NVIDIA GeForce RTX 4060 GPUs, with a batch size of 16. We conduct a total of 40 communication rounds, and each round consists of 5 epochs of local training. The learning rate is set to 0.001.

\subsection{Experimental Results}
\textbf{Style generalization. }
Table~\ref{tab:diff style} shows the contribution of $FedCD$ to style generalization. It can be observed that our method achieves an average $Acc$ increase of at least $1.45\%$ over the baselines when using both $SCI$ and $REA$, and at least $1.14\%$ when using only $SCI$. $AM$, i.e., Amplitude Mix, is a powerful Fourier-based augmentation method widely used in many $DG$ methods~\cite{wang2022domain, xu2021fourier, zhang2022semi}.
Notably, our method achieves a slightly lower $Acc$ of $0.12\%$ compared to $SNR$ when the test style is "P", which is a centralized method specifically designed to remove style confounder. However, our average $Acc$ surpasses it by $3.29\%$, with a more stable performance, as indicated by a variance that is $0.54 \times$ smaller than that of $SNR$.

$FedSR$ is a federated learning algorithm that enhances the generalization performance of federated learning through representation and conditional mutual information regularization. It can be seen that its $Acc$ is $0.66\%$ higher than our method when the test domain is "S", but its average $Acc$ is $1.29\%$ lower than ours. Our method is also more robust than it, with a variance in $Acc$ that is $0.78\times$ smaller.

\begin{table}
	\centering
	\caption{\textit{Style generalization}. Training results for accuracy $Acc$(\%) and average accuracy $AA$(\%) on four styles in the $PACS$ dataset, which are Photo (P), Art painting (A), Cartoon (C) and Sketch (S). The table header represents the current testing style, following the `leave-one-domain-out' principle, where each of the three clients corresponds to one of the remaining three styles (domains).}
	\begin{tabular}{lccccc}
		\toprule
		Algorithm  & P & A & C & S & AA\\
		\midrule
		ARFL~\cite{10234583} & 92.10 & 76.25 & 75.79 & 80.47 & 81.15\\
		FedAvg~\cite{pmlr-v54-mcmahan17a} & 92.77 & 77.29 & 77.97 & 81.03 & 82.27\\
		FedDG-GA~\cite{zhang2023federated} & 93.97 & 81.28 & 76.73 & 82.57 & 83.64\\
		FedCSA~\cite{dong2023fed} & 91.88 & 77.00 & 76.79 & 80.84 & 81.63\\
		FedNova~\cite{NEURIPS2020_564127c0} & 94.03 & 79.93 & 76.39 & 79.26 & 82.40\\
		FedProx~\cite{li2020federated} & 93.15 & 77.72 & 77.73 & 80.77 & 82.34\\
		FedSAM~\cite{qu2022generalized} & 91.20 & 74.45 & 77.77 & 83.35 & 81.69\\
		FedADG~\cite{10011632} & 92.93 & 77.85 & 74.74 & 79.54 & 81.27\\
		HarmoFL~\cite{jiang2022harmofl} & 90.99 & 74.51 & 77.43 & 81.73 & 81.17\\
		Scaffold~\cite{karimireddy2020scaffold} & 92.50 & 78.09 & 77.23 & 80.67 & 82.12\\
		FedSR~\cite{NEURIPS2022_fd946a6c} & 93.82 & \textbf{83.24} & 76.03 & 82.11 & 83.80\\
		FedCMI & 92.85 & 80.84 & 73.72 & 79.52 & 81.73\\
		FedL2R & 92.84 & 82.24 & 75.83 & 81.61 & 83.13\\
		AM & 93.29 & 80.86 & 77.62 & 81.05 & 83.21\\
		RSC~\cite{huang2020self} & 92.67 & 77.98 & 77.80 & 82.90 & 82.84\\
		SNR~\cite{jin2021style} & \textbf{94.54} & 80.32 & 78.23 & 74.12 & 81.80\\
		\midrule
		FedCD$-SCI$ & 94.13 & 82.27 & 79.05 & 83.66 & 84.78\\
		\midrule
		FedCD$-SCI+REA$ & 94.42 & 82.58 & \textbf{79.35} & \textbf{84.00} & \textbf{85.09} \\
		\bottomrule
	\end{tabular}
	\label{tab:diff style}
\end{table}

\begin{table}
	\centering
	\caption{\textit{Dataset generalization}. Training results for $mAP_{50}(\%)$ and average $mAP_{50}$ \textit{Avg}(\%) in three datasets (domains) in Cityscapes (C), Mapillary (M) and BDD100K (B), with "leave-one-domain-out".}
	\begin{tabular}{lcccc}
		\toprule
		Algorithm  & C & M & B & \textit{Avg}\\
		\midrule
		FedAvg     & 43.39  & 35.49 & 39.69 & 39.52\\
		FedProx  & 44.10 & 35.56 & 40.10 & 39.92\\
		FedSAM & 43.61 & 33.77 & 44.14 & 40.51\\
		FedDG-GA & 43.75 & 35.49 & 40.94 & 40.06\\
		\midrule
		\makecell[l]{FedCD$-SCI$} & 46.47 & 40.00 & 44.50 & 43.65\\
		\midrule
		\makecell[l]{FedCD$-SCI+REA$} & \textbf{46.67} & \textbf{41.37} & \textbf{46.56} & \textbf{44.86} \\
		\bottomrule
	\end{tabular}
	\label{tab:diff dataset}
\end{table}
\begin{table}
	\centering
	\caption{\textit{Scene generalization}. Training results for $mAP_{50}(\%)$ and average $mAP_{50}$ \textit{Avg}(\%) in four scenes (domains), ClearHighway (CH), RainyHighway (RH), CityStreet (CS), Residential (RE),  within $BDD100K$ dataset, with the "leave-one-domain-out" principle.}
	\begin{tabular}{lccccc}
		\toprule
		Algorithm  & CH & RH & CS & RE & \textit{Avg}\\
		\midrule
		FedAvg     & 37.98  & 38.26 & 41.91 & 45.21 & 40.84\\
		FedProx  & 38.69 & 38.89 & 41.82 & 44.46 & 40.97\\
		FedSAM & 38.44 & 39.77 & 40.31 & 46.04 & 41.14\\
		FedDG-GA & 39.05 & 39.74 & 41.86 & 45.91 & 41.64\\
		\midrule
		\makecell[l]{FedCD$-SCI$} & 41.44 & 38.81 & 41.21 & 46.58 & 42.26\\
		\midrule
		\makecell[l]{FedCD$-SCI+REA$} & \textbf{41.69} & \textbf{39.78} & \textbf{43.11} & \textbf{47.05} & \textbf{42.91} \\
		\bottomrule
	\end{tabular}
	\label{tab:diff scene}
\end{table}
\textbf{Dataset/Scene generalization. }
In the dataset generalization setting shown in Table~\ref{tab:diff dataset}, $FedCD$ achieves an average $mAP_{50}$ increase of at least $4.8\%$ over baselines when using both $SCI$ and $REA$, and at least $3.14\%$ when using only $SCI$. In the scene generalization setting shown in Table~\ref{tab:diff scene}, $FedCD$ achieves an average $mAP_{50}$ increase of at least $1.27\%$ over baselines when using both $SCI$ and $REA$, and at least $0.62\%$ when using only $SCI$. Both of these scenarios demonstrate very small variances of $6.12 (\%^2)$ and $7.12 (\%^2)$, respectively, indicating the stability of our method's performance.

It is worth noting that our approach yields greater performance improvements in the dataset generalization setting with larger domain shifts. We attribute this to the inherent generalization capability of neural networks, which can to some extent address stable predictions with smaller domain shifts. Therefore, the improvement brought by our approach appears relatively small.
\subsection{Analyses on $SCI$}\label{Analyses on IMG}
\begin{figure}[htbp]
	\centering
	\begin{subfigure}[b]{0.23\textwidth}
		\centering
		\includegraphics[width=\textwidth]{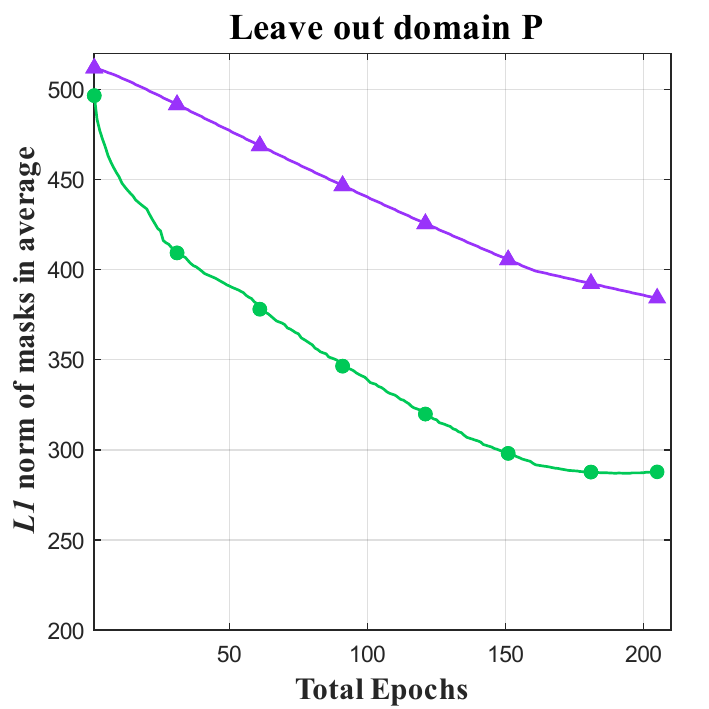}
		\caption{Leave out P}
		\label{fig:sub1}
	\end{subfigure}
	\begin{subfigure}[b]{0.23\textwidth}
		\centering
		\includegraphics[width=\textwidth]{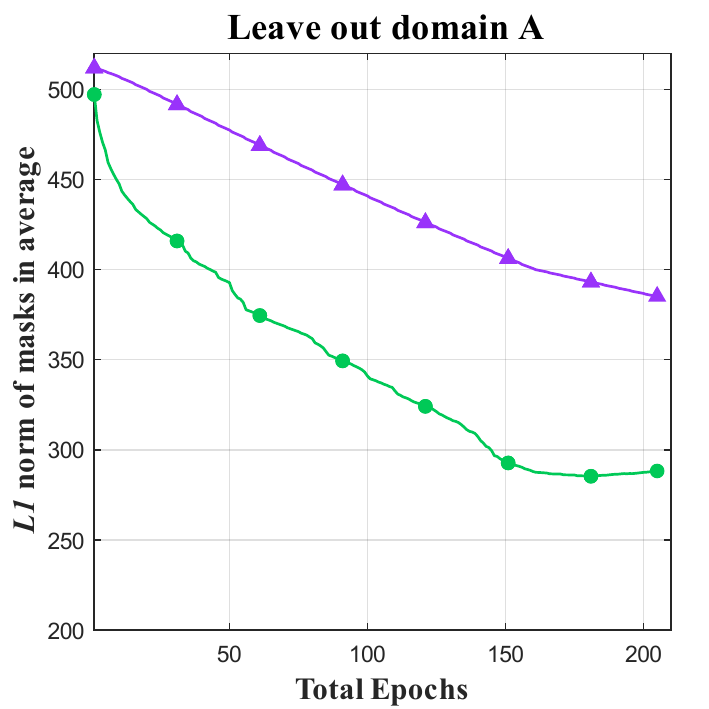}
		\caption{Leave out A}
		\label{fig:sub2}
	\end{subfigure}
	
	\begin{subfigure}[b]{0.23\textwidth}
		\centering
		\includegraphics[width=\textwidth]{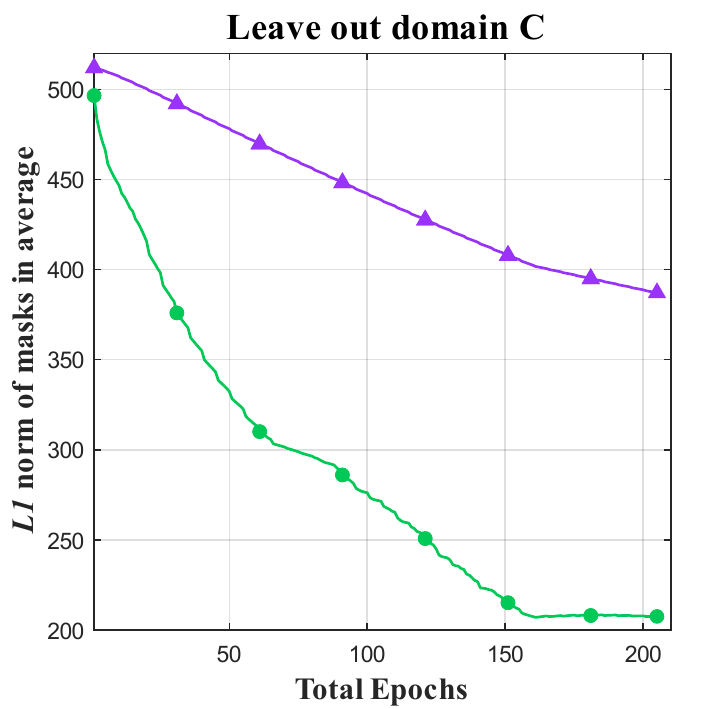}
		\caption{Leave out C}
		\label{fig:sub3}
	\end{subfigure}
	\begin{subfigure}[b]{0.23\textwidth}
		\centering
		\includegraphics[width=\textwidth]{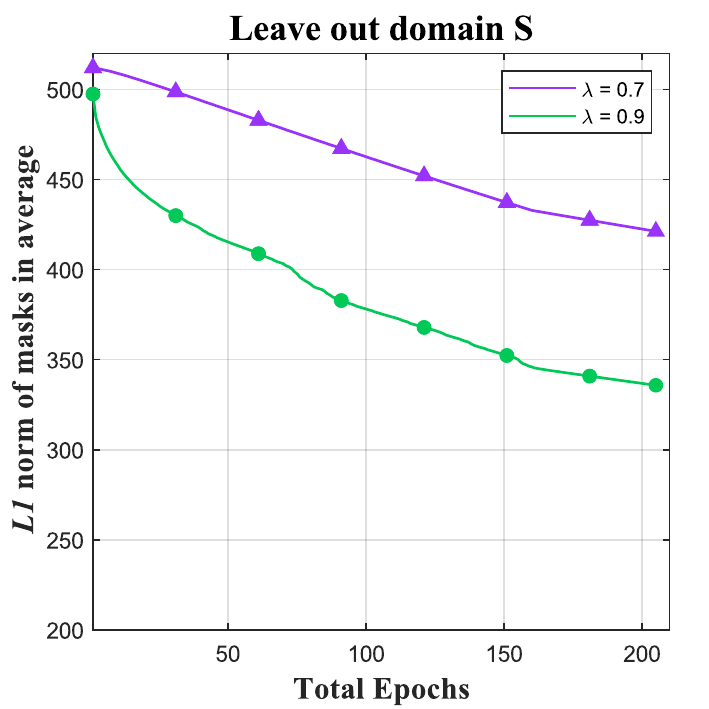}
		\caption{Leave out S}
		\label{fig:sub4}
	\end{subfigure}
	\caption{Comparing the average \textit{L1} norm of the masks generated by $SCI$ under different hyperparameters $\lambda = 0.7/0.9$ and different "leave-one-domain-out" settings.}\label{fig:IMG results}
\end{figure}

\textbf{Sparsity analysis. } 
We continuously observed the \textit{L1} norm of the average mask on randomly sampled batches during training. For more details, we conducted experiments under each "leave-one-domain-out" setting with $\lambda = \{0.7, 0.9\}$ respectively, and the results are shown in 
Figure~\ref{fig:IMG results}.
We calculated the \textit{L1} norm of all masks generated on each participating client. Subsequently, we averaged these norms both within each client and across all clients. This process allowed us to observe the trend of the \textit{L1} norm with the total training epochs of federated learning.

We can observe from Figure~\ref{fig:IMG results} that, the $L1$ norm of the mask shows a decreasing trend for different settings of $\lambda$ and leave-out domain. Since the $L1$ norm is essentially a sparse operator, this indicates that features tend to become sparse during the federated training.
This is actually consistent with the insight provided by $FedL2R$ in~\cite{NEURIPS2022_fd946a6c}, which directly utilizes the regularization term:
\begin{align}
	l^{L2R} = \mathbb{E}_{p_i(z)}[||z||^2_2]\label{fedl2r regularization},
\end{align}
where $z$ represents the intermediate feature representation of the network. Using the same notation, we can consider $z = M_{\delta_e}(x) = M_e \odot x$, where for a sparser $M_e$, a sparser $z$ can be generated, resulting in a smaller $l_{L2R}$.

Assuming there exists a "reference" distribution $q(z) = \mathcal{N}(0, \sigma^2I)$ (with a small $\sigma$), we have: 
\begin{align}
	-logq(z) = \frac{||z||^2_2}{2\sigma^2} + C, \label{reference distribution}
\end{align}
where \textit{C} is an addictive constant.

We can consider incorporating $q(z)$ into the calculation of $l^{L2R}$:
\begin{equation}
	\begin{aligned}
		&l^{L2R} = \mathbb{E}_{p_i(z)}[||z||^2_2]
		=2\sigma^2\mathbb{E}_{p_i(z)}[-logq(z)]
		=2\sigma^2H(p_i(z), q(z)),
		\label{cross entropy form}
	\end{aligned}
\end{equation}
where $H(p_i(z), q(z))$ denotes the cross entropy from $p_i(z)$ to the reference distribution $q(z)$.

The Eq.~\ref{cross entropy form} can be rewritten in the form of KL divergence:
\begin{equation}
	\begin{aligned}
		H(p_i(z), q(z)) = H(p_i(z)) + KL[p_i(z) || q(z)].
		\label{dKL form}
	\end{aligned}
\end{equation}
\begin{figure*} 
	\centering
	\begin{subfigure}[b]{0.47\textwidth}
		\centering
		\includegraphics[width=\textwidth]{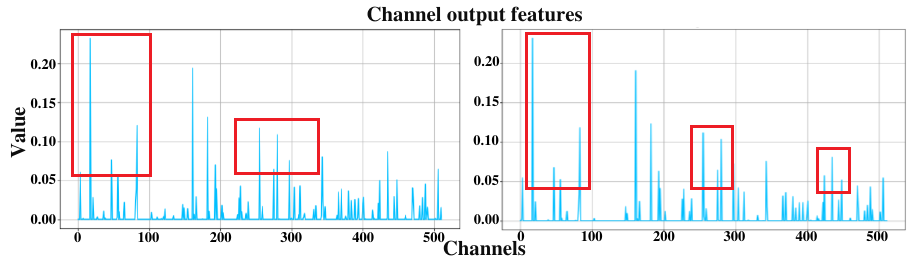}
		\caption{Features in A}
		\label{feature_71_a}
	\end{subfigure}
	\begin{subfigure}[b]{0.47\textwidth}
		\centering
		\includegraphics[width=\textwidth]{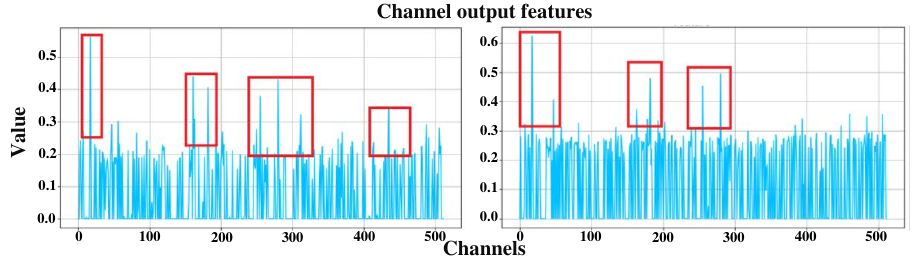}
		\caption{Features in S}
		\label{feature_66_s}
	\end{subfigure}

	\begin{subfigure}[b]{0.47\textwidth}
		\centering
		\includegraphics[width=\textwidth]{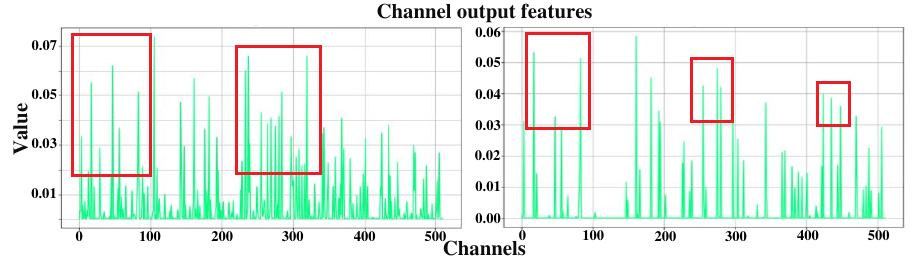}
		\caption{Masked Features in A}
		\label{feature_mask_71_a}
	\end{subfigure}
	\begin{subfigure}[b]{0.47\textwidth}
		\centering
		\includegraphics[width=\textwidth]{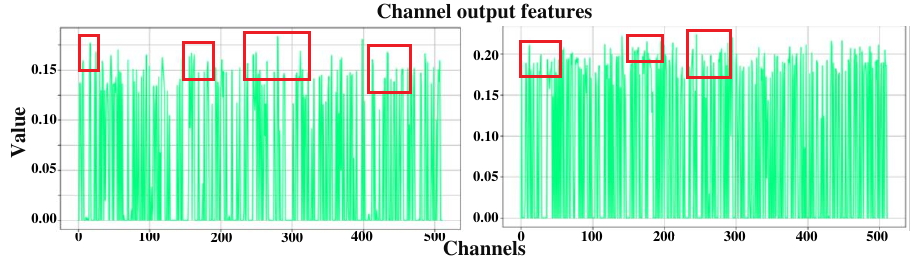}
		\caption{Masked Features in S}
		\label{feature_mask_66_s}
	\end{subfigure}
	
	\begin{subfigure}[b]{0.47\textwidth}
		\centering
		\includegraphics[width=\textwidth]{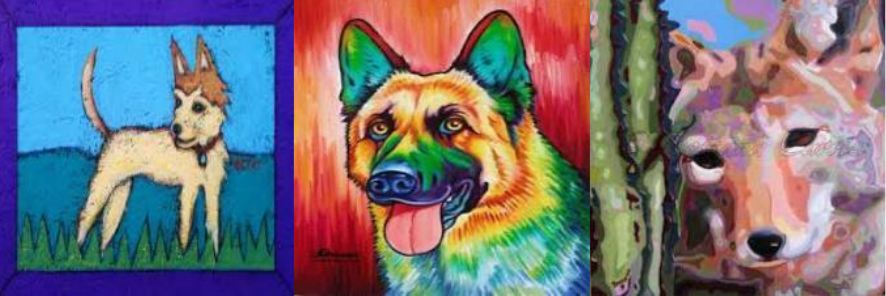}
		\caption{Samples in A}
		\label{artpainting_raw_img}
	\end{subfigure}
	\begin{subfigure}[b]{0.47\textwidth}
		\centering
		\includegraphics[width=\textwidth]{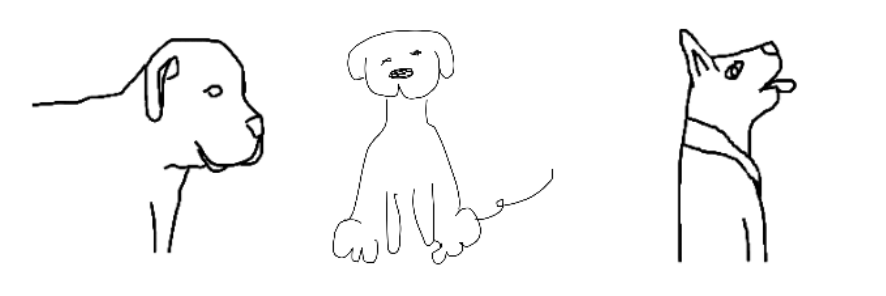}
		\caption{Samples in S}
		\label{sketch_raw_img}
	\end{subfigure}
	\caption{The comparison between the original features and masked features of samples from feature-rich domain Art painting (\textbf{A}) and feature-poor domain Sketch (\textbf{S}), which reflects two phenomena: a decrease in overall and a decrease in specificity.}\label{fig:mask_on_artpainting}
\end{figure*}
Therefore, if the entropy $H(p_i(z))$ remains relatively stable during training, minimizing $\ell^{L2R}$ will also minimize $KL[p_i(z)||q(z)]$, which encourages $p_i(z)$ to be close to $q(z)$, implying an implicit alignment of the marginal distribution. Therefore, our method also aligns the feature distributions of each client to some extent, establishing a connection with domain-invariant representation learning, thereby improving generalization. However, our method achieves an $Acc$ $1.65$ higher than $FedL2R$ when using only $SCI$, which reflects the gain from considering the additional spurious correlations.

\subsection{DISCUSSION}
\textbf{Masked feature analysis. }Next, we analyzed the feature outputs before and after adding the mask. In the experiment setting of leaving out "P", we took the \textit{art painting} style with rich features and the \textit{sketch} style with fewer features as examples, as shown in 
Figure~\ref{fig:mask_on_artpainting}. Because we added a mask to the output of the $Avg\_pool$ layer of $ResNet18$, which is channel-wise average pooling, we obtained a one-dimensional tensor, and our mask equivalently acts on all the feature pixels before pooling.

For both styles with rich features and styles with singular features, our masks exhibit two trends: a decrease in overall and a decrease in specificity.

The decrease in overall may correspond to some intra-channel overall spurious correlated features, such as background information, style information, etc., which exist in the feature maps of each channel. This actually corresponds to the conclusion in Section~\ref{Analyses on IMG}. As shown in Figure~\ref{fig:mask_on_artpainting}, the majority of original image feature values are 0.05 and 0.2 in domain "A" and "S". After adding masks, they are reduced to 0.03 and 0.15 respectively, with reduction factors of $1.7\times$ and $1.3\times$. This indicates that for images like the ones in Figure~\ref{artpainting_raw_img} in domain "A", which may have more overall spurious shortcut information in each channel based on their own feature richness, we would produce an unbiasedly larger weakening compared to samples that already have insufficiently rich information, like the ones in Figure~\ref{sketch_raw_img} in domain "S". 

Regarding the decrease in specificity, for the feature values highlighted by the red boxes in Figure~\ref{feature_71_a} and Figure~\ref{feature_66_s}, our mask can generate specific weakening, i.e., changes in relative size. This corresponds to the extraction of more spurious correlated features in certain channels. Furthermore, it can be observed that the masked feature in Figure~\ref{feature_mask_66_s} has a more uniform channel-wise distribution compared to Figure~\ref{feature_mask_71_a}. We believe this is because the features of domain "S" samples are singular, leading to more similar causally invariant features in each channel and thus forming a more uniform feature distribution.

\section{Conclusion}

We primarily address the domain generalization issue in multimedia federated learning and propose a simple yet effective framework, $FedCD$, that reconsiders this problem from both local and global perspectives. In local level, our $SCI$ differs from previous approaches, which considered all features unbiasedly during training. Instead, we generate feature interveners in a self-supervised manner based on the principle of invariance. This approach helps intervene and reduce the adverse effects of spurious shortcut features, all without requiring data or feature-related information sharing. Additionally, it does not rely on any prior knowledge of causal invariance components in the images, further expanding the scope of removing spurious features due to its nonspecific purpose. We provide theoretical derivations for our approach. On the server side, we introduce a novel aggregation module, $REA$, which utilizes $ERM$ to achieve better global causal invariant predictions by converting the aggregation coefficients into the solution of a mathematical optimization problem. This process can be solved by simply using sequential least squares without introducing new neural networks. Additionally, we conduct thorough ablation experiments and analysis to validate the effectiveness of our method in various tasks and generalization scenarios, demonstrating its robustness and scalability in complex multimedia datasets.


\bibliographystyle{ACM_Reference_Format}
\bibliography{sample_base}


\begin{thebibliography}{50}


\ifx \showCODEN    \undefined \def \showCODEN     #1{\unskip}     \fi
\ifx \showDOI      \undefined \def \showDOI       #1{#1}\fi
\ifx \showISBNx    \undefined \def \showISBNx     #1{\unskip}     \fi
\ifx \showISBNxiii \undefined \def \showISBNxiii  #1{\unskip}     \fi
\ifx \showISSN     \undefined \def \showISSN      #1{\unskip}     \fi
\ifx \showLCCN     \undefined \def \showLCCN      #1{\unskip}     \fi
\ifx \shownote     \undefined \def \shownote      #1{#1}          \fi
\ifx \showarticletitle \undefined \def \showarticletitle #1{#1}   \fi
\ifx \showURL      \undefined \def \showURL       {\relax}        \fi
\providecommand\bibfield[2]{#2}
\providecommand\bibinfo[2]{#2}
\providecommand\natexlab[1]{#1}
\providecommand\showeprint[2][]{arXiv:#2}

\bibitem[Antequera et~al\mbox{.}(2020)]%
        {antequera2020mapillary}
\bibfield{author}{\bibinfo{person}{Manuel~L{\'o}pez Antequera}, \bibinfo{person}{Pau Gargallo}, \bibinfo{person}{Markus Hofinger}, \bibinfo{person}{Samuel~Rota Bul{\`o}}, \bibinfo{person}{Yubin Kuang}, {and} \bibinfo{person}{Peter Kontschieder}.} \bibinfo{year}{2020}\natexlab{}.
\newblock \showarticletitle{Mapillary planet-scale depth dataset}. In \bibinfo{booktitle}{\emph{the European Conference on Computer Vision}}. Springer, \bibinfo{pages}{589--604}.
\newblock


\bibitem[Arjovsky et~al\mbox{.}(2019)]%
        {arjovsky2019invariant}
\bibfield{author}{\bibinfo{person}{Martin Arjovsky}, \bibinfo{person}{L{\'e}on Bottou}, \bibinfo{person}{Ishaan Gulrajani}, {and} \bibinfo{person}{David Lopez-Paz}.} \bibinfo{year}{2019}\natexlab{}.
\newblock \showarticletitle{Invariant risk minimization}.
\newblock \bibinfo{journal}{\emph{arXiv preprint arXiv:1907.02893}} (\bibinfo{year}{2019}).
\newblock


\bibitem[Balaji et~al\mbox{.}(2018)]%
        {NEURIPS2018_647bba34}
\bibfield{author}{\bibinfo{person}{Yogesh Balaji}, \bibinfo{person}{Swami Sankaranarayanan}, {and} \bibinfo{person}{Rama Chellappa}.} \bibinfo{year}{2018}\natexlab{}.
\newblock \showarticletitle{MetaReg: Towards Domain Generalization using Meta-Regularization}. In \bibinfo{booktitle}{\emph{Advances in Neural Information Processing Systems}}, Vol.~\bibinfo{volume}{31}.
\newblock


\bibitem[Chen et~al\mbox{.}(2023)]%
        {Chen_2023_WACV}
\bibfield{author}{\bibinfo{person}{Junming Chen}, \bibinfo{person}{Meirui Jiang}, \bibinfo{person}{Qi Dou}, {and} \bibinfo{person}{Qifeng Chen}.} \bibinfo{year}{2023}\natexlab{}.
\newblock \showarticletitle{Federated Domain Generalization for Image Recognition via Cross-Client Style Transfer}. In \bibinfo{booktitle}{\emph{the IEEE/CVF Winter Conference on Applications of Computer Vision}}. \bibinfo{pages}{361--370}.
\newblock


\bibitem[Cordts et~al\mbox{.}(2016)]%
        {cordts2016cityscapes}
\bibfield{author}{\bibinfo{person}{Marius Cordts}, \bibinfo{person}{Mohamed Omran}, \bibinfo{person}{Sebastian Ramos}, \bibinfo{person}{Timo Rehfeld}, \bibinfo{person}{Markus Enzweiler}, \bibinfo{person}{Rodrigo Benenson}, \bibinfo{person}{Uwe Franke}, \bibinfo{person}{Stefan Roth}, {and} \bibinfo{person}{Bernt Schiele}.} \bibinfo{year}{2016}\natexlab{}.
\newblock \showarticletitle{The cityscapes dataset for semantic urban scene understanding}. In \bibinfo{booktitle}{\emph{the IEEE/CVF International Conference on Computer Vision}}. \bibinfo{pages}{3213--3223}.
\newblock


\bibitem[Dong et~al\mbox{.}(2023)]%
        {dong2023fed}
\bibfield{author}{\bibinfo{person}{Xinyu Dong}, \bibinfo{person}{Zhenwei Shi}, \bibinfo{person}{XiaoMei Huang}, \bibinfo{person}{Chu Han}, \bibinfo{person}{Zihan Cao}, \bibinfo{person}{Zhihe Zhao}, \bibinfo{person}{Dan Wang}, \bibinfo{person}{Peng Xu}, \bibinfo{person}{Zaiyi Liu}, {and} \bibinfo{person}{Wenbin Liu}.} \bibinfo{year}{2023}\natexlab{}.
\newblock \showarticletitle{Fed-CSA: Channel Spatial Attention and Adaptive Weights Aggregation-Based Federated Learning for Breast Tumor Segmentation on MRI}. In \bibinfo{booktitle}{\emph{International Conference on Intelligent Computing}}. Springer, \bibinfo{pages}{312--323}.
\newblock


\bibitem[Dou et~al\mbox{.}(2019)]%
        {NEURIPS2019_2974788b}
\bibfield{author}{\bibinfo{person}{Qi Dou}, \bibinfo{person}{Daniel Coelho~de Castro}, \bibinfo{person}{Konstantinos Kamnitsas}, {and} \bibinfo{person}{Ben Glocker}.} \bibinfo{year}{2019}\natexlab{}.
\newblock \showarticletitle{Domain Generalization via Model-Agnostic Learning of Semantic Features}. In \bibinfo{booktitle}{\emph{Advances in Neural Information Processing Systems}}, Vol.~\bibinfo{volume}{32}.
\newblock


\bibitem[He et~al\mbox{.}(2016)]%
        {He_2016_CVPR}
\bibfield{author}{\bibinfo{person}{Kaiming He}, \bibinfo{person}{Xiangyu Zhang}, \bibinfo{person}{Shaoqing Ren}, {and} \bibinfo{person}{Jian Sun}.} \bibinfo{year}{2016}\natexlab{}.
\newblock \showarticletitle{Deep Residual Learning for Image Recognition}. In \bibinfo{booktitle}{\emph{the IEEE/CVF International Conference on Computer Vision}}.
\newblock


\bibitem[Huang et~al\mbox{.}(2020)]%
        {huang2020self}
\bibfield{author}{\bibinfo{person}{Zeyi Huang}, \bibinfo{person}{Haohan Wang}, \bibinfo{person}{Eric~P Xing}, {and} \bibinfo{person}{Dong Huang}.} \bibinfo{year}{2020}\natexlab{}.
\newblock \showarticletitle{Self-challenging improves cross-domain generalization}. In \bibinfo{booktitle}{\emph{the European Conference on Computer Vision}}. Springer, \bibinfo{pages}{124--140}.
\newblock


\bibitem[Jiang et~al\mbox{.}(2022)]%
        {jiang2022harmofl}
\bibfield{author}{\bibinfo{person}{Meirui Jiang}, \bibinfo{person}{Zirui Wang}, {and} \bibinfo{person}{Qi Dou}.} \bibinfo{year}{2022}\natexlab{}.
\newblock \showarticletitle{Harmofl: Harmonizing local and global drifts in federated learning on heterogeneous medical images}. In \bibinfo{booktitle}{\emph{the AAAI Conference on Artificial Intelligence}}, Vol.~\bibinfo{volume}{36}. \bibinfo{pages}{1087--1095}.
\newblock


\bibitem[Jin et~al\mbox{.}(2021)]%
        {jin2021style}
\bibfield{author}{\bibinfo{person}{Xin Jin}, \bibinfo{person}{Cuiling Lan}, \bibinfo{person}{Wenjun Zeng}, {and} \bibinfo{person}{Zhibo Chen}.} \bibinfo{year}{2021}\natexlab{}.
\newblock \showarticletitle{Style normalization and restitution for domain generalization and adaptation}.
\newblock \bibinfo{journal}{\emph{IEEE Transactions on Multimedia}}  \bibinfo{volume}{24} (\bibinfo{year}{2021}), \bibinfo{pages}{3636--3651}.
\newblock


\bibitem[Karimireddy et~al\mbox{.}(2020)]%
        {karimireddy2020scaffold}
\bibfield{author}{\bibinfo{person}{Sai~Praneeth Karimireddy}, \bibinfo{person}{Satyen Kale}, \bibinfo{person}{Mehryar Mohri}, \bibinfo{person}{Sashank Reddi}, \bibinfo{person}{Sebastian Stich}, {and} \bibinfo{person}{Ananda~Theertha Suresh}.} \bibinfo{year}{2020}\natexlab{}.
\newblock \showarticletitle{Scaffold: Stochastic controlled averaging for federated learning}. In \bibinfo{booktitle}{\emph{International Conference on Machine Learning}}. PMLR, \bibinfo{pages}{5132--5143}.
\newblock


\bibitem[Koyama and Yamaguchi(2021)]%
        {koyama2021invariance}
\bibfield{author}{\bibinfo{person}{Masanori Koyama} {and} \bibinfo{person}{Shoichiro Yamaguchi}.} \bibinfo{year}{2021}\natexlab{}.
\newblock \bibinfo{title}{When is invariance useful in an Out-of-Distribution Generalization problem ?}
\newblock
\newblock
\showeprint[arxiv]{2008.01883}


\bibitem[Krueger et~al\mbox{.}(2021)]%
        {krueger2021out}
\bibfield{author}{\bibinfo{person}{David Krueger}, \bibinfo{person}{Ethan Caballero}, \bibinfo{person}{Joern-Henrik Jacobsen}, \bibinfo{person}{Amy Zhang}, \bibinfo{person}{Jonathan Binas}, \bibinfo{person}{Dinghuai Zhang}, \bibinfo{person}{Remi Le~Priol}, {and} \bibinfo{person}{Aaron Courville}.} \bibinfo{year}{2021}\natexlab{}.
\newblock \showarticletitle{Out-of-distribution generalization via risk extrapolation (rex)}. In \bibinfo{booktitle}{\emph{International Conference on Machine Learning}}. PMLR, \bibinfo{pages}{5815--5826}.
\newblock


\bibitem[Li et~al\mbox{.}(2018c)]%
        {li2018learning}
\bibfield{author}{\bibinfo{person}{Da Li}, \bibinfo{person}{Yongxin Yang}, \bibinfo{person}{Yi-Zhe Song}, {and} \bibinfo{person}{Timothy Hospedales}.} \bibinfo{year}{2018}\natexlab{c}.
\newblock \showarticletitle{Learning to generalize: Meta-learning for domain generalization}. In \bibinfo{booktitle}{\emph{the AAAI Conference on Artificial Intelligence}}, Vol.~\bibinfo{volume}{32}.
\newblock


\bibitem[Li et~al\mbox{.}(2017)]%
        {li2017deeper}
\bibfield{author}{\bibinfo{person}{Da Li}, \bibinfo{person}{Yongxin Yang}, \bibinfo{person}{Yi-Zhe Song}, {and} \bibinfo{person}{Timothy~M Hospedales}.} \bibinfo{year}{2017}\natexlab{}.
\newblock \showarticletitle{Deeper, broader and artier domain generalization}. In \bibinfo{booktitle}{\emph{the IEEE/CVF International Conference on Computer Vision}}. \bibinfo{pages}{5542--5550}.
\newblock


\bibitem[Li et~al\mbox{.}(2018a)]%
        {li2018domain}
\bibfield{author}{\bibinfo{person}{Haoliang Li}, \bibinfo{person}{Sinno~Jialin Pan}, \bibinfo{person}{Shiqi Wang}, {and} \bibinfo{person}{Alex~C Kot}.} \bibinfo{year}{2018}\natexlab{a}.
\newblock \showarticletitle{Domain generalization with adversarial feature learning}. In \bibinfo{booktitle}{\emph{the IEEE/CVF Conference on Computer Vision and Pattern Recognition}}. \bibinfo{pages}{5400--5409}.
\newblock


\bibitem[Li et~al\mbox{.}(2022)]%
        {li2022federated}
\bibfield{author}{\bibinfo{person}{Qinbin Li}, \bibinfo{person}{Yiqun Diao}, \bibinfo{person}{Quan Chen}, {and} \bibinfo{person}{Bingsheng He}.} \bibinfo{year}{2022}\natexlab{}.
\newblock \showarticletitle{Federated learning on non-iid data silos: An experimental study}. In \bibinfo{booktitle}{\emph{International Conference on Data Engineering}}. IEEE, \bibinfo{pages}{965--978}.
\newblock


\bibitem[Li et~al\mbox{.}(2021)]%
        {li2021model}
\bibfield{author}{\bibinfo{person}{Qinbin Li}, \bibinfo{person}{Bingsheng He}, {and} \bibinfo{person}{Dawn Song}.} \bibinfo{year}{2021}\natexlab{}.
\newblock \showarticletitle{Model-contrastive federated learning}. In \bibinfo{booktitle}{\emph{the IEEE/CVF Conference on Computer Vision and Pattern Recognition}}. \bibinfo{pages}{10713--10722}.
\newblock


\bibitem[Li et~al\mbox{.}(2020)]%
        {li2020federated}
\bibfield{author}{\bibinfo{person}{Tian Li}, \bibinfo{person}{Anit~Kumar Sahu}, \bibinfo{person}{Manzil Zaheer}, \bibinfo{person}{Maziar Sanjabi}, \bibinfo{person}{Ameet Talwalkar}, {and} \bibinfo{person}{Virginia Smith}.} \bibinfo{year}{2020}\natexlab{}.
\newblock \showarticletitle{Federated optimization in heterogeneous networks}. In \bibinfo{booktitle}{\emph{Machine Learning and Systems}}, Vol.~\bibinfo{volume}{2}. \bibinfo{pages}{429--450}.
\newblock


\bibitem[Li et~al\mbox{.}(2018b)]%
        {li2018deep}
\bibfield{author}{\bibinfo{person}{Ya Li}, \bibinfo{person}{Xinmei Tian}, \bibinfo{person}{Mingming Gong}, \bibinfo{person}{Yajing Liu}, \bibinfo{person}{Tongliang Liu}, \bibinfo{person}{Kun Zhang}, {and} \bibinfo{person}{Dacheng Tao}.} \bibinfo{year}{2018}\natexlab{b}.
\newblock \showarticletitle{Deep domain generalization via conditional invariant adversarial networks}. In \bibinfo{booktitle}{\emph{the European Conference on Computer Vision}}. \bibinfo{pages}{624--639}.
\newblock


\bibitem[Liu et~al\mbox{.}(2021)]%
        {liu2021feddg}
\bibfield{author}{\bibinfo{person}{Quande Liu}, \bibinfo{person}{Cheng Chen}, \bibinfo{person}{Jing Qin}, \bibinfo{person}{Qi Dou}, {and} \bibinfo{person}{Pheng-Ann Heng}.} \bibinfo{year}{2021}\natexlab{}.
\newblock \showarticletitle{Fed{DG}: Federated domain generalization on medical image segmentation via episodic learning in continuous frequency space}. In \bibinfo{booktitle}{\emph{the IEEE/CVF Conference on Computer Vision and Pattern Recognition}}. \bibinfo{pages}{1013--1023}.
\newblock


\bibitem[Lv et~al\mbox{.}(2022)]%
        {lv2022causality}
\bibfield{author}{\bibinfo{person}{Fangrui Lv}, \bibinfo{person}{Jian Liang}, \bibinfo{person}{Shuang Li}, \bibinfo{person}{Bin Zang}, \bibinfo{person}{Chi~Harold Liu}, \bibinfo{person}{Ziteng Wang}, {and} \bibinfo{person}{Di Liu}.} \bibinfo{year}{2022}\natexlab{}.
\newblock \showarticletitle{Causality inspired representation learning for domain generalization}. In \bibinfo{booktitle}{\emph{the IEEE/CVF Conference on Computer Vision and Pattern Recognition}}. \bibinfo{pages}{8046--8056}.
\newblock


\bibitem[McMahan et~al\mbox{.}(2017)]%
        {pmlr-v54-mcmahan17a}
\bibfield{author}{\bibinfo{person}{Brendan McMahan}, \bibinfo{person}{Eider Moore}, \bibinfo{person}{Daniel Ramage}, \bibinfo{person}{Seth Hampson}, {and} \bibinfo{person}{Blaise Aguera~y Arcas}.} \bibinfo{year}{2017}\natexlab{}.
\newblock \showarticletitle{Communication-Efficient Learning of Deep Networks from Decentralized Data}. In \bibinfo{booktitle}{\emph{Conference on Artificial Intelligence and Statistics}}. \bibinfo{publisher}{PMLR}, \bibinfo{address}{Ft. Lauderdale, FL}, \bibinfo{pages}{1273--1282}.
\newblock


\bibitem[Mu et~al\mbox{.}(2023)]%
        {mu2023fedproc}
\bibfield{author}{\bibinfo{person}{Xutong Mu}, \bibinfo{person}{Yulong Shen}, \bibinfo{person}{Ke Cheng}, \bibinfo{person}{Xueli Geng}, \bibinfo{person}{Jiaxuan Fu}, \bibinfo{person}{Tao Zhang}, {and} \bibinfo{person}{Zhiwei Zhang}.} \bibinfo{year}{2023}\natexlab{}.
\newblock \showarticletitle{Fedproc: Prototypical contrastive federated learning on non-iid data}.
\newblock \bibinfo{journal}{\emph{Future Generation Computer Systems}}  \bibinfo{volume}{143} (\bibinfo{year}{2023}), \bibinfo{pages}{93--104}.
\newblock


\bibitem[Muandet et~al\mbox{.}(2013)]%
        {muandet2013domain}
\bibfield{author}{\bibinfo{person}{Krikamol Muandet}, \bibinfo{person}{David Balduzzi}, {and} \bibinfo{person}{Bernhard Sch{\"o}lkopf}.} \bibinfo{year}{2013}\natexlab{}.
\newblock \showarticletitle{Domain generalization via invariant feature representation}. In \bibinfo{booktitle}{\emph{International Conference on Machine Learning}}. PMLR, \bibinfo{pages}{10--18}.
\newblock


\bibitem[Nguyen et~al\mbox{.}(2022)]%
        {NEURIPS2022_fd946a6c}
\bibfield{author}{\bibinfo{person}{A.~Tuan Nguyen}, \bibinfo{person}{Philip Torr}, {and} \bibinfo{person}{Ser~Nam Lim}.} \bibinfo{year}{2022}\natexlab{}.
\newblock \showarticletitle{FedSR: A Simple and Effective Domain Generalization Method for Federated Learning}. In \bibinfo{booktitle}{\emph{Advances in Neural Information Processing Systems}}, \bibfield{editor}{\bibinfo{person}{S.~Koyejo}, \bibinfo{person}{S.~Mohamed}, \bibinfo{person}{A.~Agarwal}, \bibinfo{person}{D.~Belgrave}, \bibinfo{person}{K.~Cho}, {and} \bibinfo{person}{A.~Oh}} (Eds.), Vol.~\bibinfo{volume}{35}. \bibinfo{pages}{38831--38843}.
\newblock


\bibitem[Qu et~al\mbox{.}(2022)]%
        {qu2022generalized}
\bibfield{author}{\bibinfo{person}{Zhe Qu}, \bibinfo{person}{Xingyu Li}, \bibinfo{person}{Rui Duan}, \bibinfo{person}{Yao Liu}, \bibinfo{person}{Bo Tang}, {and} \bibinfo{person}{Zhuo Lu}.} \bibinfo{year}{2022}\natexlab{}.
\newblock \showarticletitle{Generalized federated learning via sharpness aware minimization}. In \bibinfo{booktitle}{\emph{International Conference on Machine Learning}}. PMLR, \bibinfo{pages}{18250--18280}.
\newblock


\bibitem[Shao et~al\mbox{.}(2021)]%
        {shao2021improving}
\bibfield{author}{\bibinfo{person}{Feifei Shao}, \bibinfo{person}{Yawei Luo}, \bibinfo{person}{Li Zhang}, \bibinfo{person}{Lu Ye}, \bibinfo{person}{Siliang Tang}, \bibinfo{person}{Yi Yang}, {and} \bibinfo{person}{Jun Xiao}.} \bibinfo{year}{2021}\natexlab{}.
\newblock \showarticletitle{Improving weakly supervised object localization via causal intervention}. In \bibinfo{booktitle}{\emph{the ACM International Conference on Multimedia}}. \bibinfo{pages}{3321--3329}.
\newblock


\bibitem[Shao et~al\mbox{.}(2019)]%
        {shao2019multi}
\bibfield{author}{\bibinfo{person}{Rui Shao}, \bibinfo{person}{Xiangyuan Lan}, \bibinfo{person}{Jiawei Li}, {and} \bibinfo{person}{Pong~C Yuen}.} \bibinfo{year}{2019}\natexlab{}.
\newblock \showarticletitle{Multi-adversarial discriminative deep domain generalization for face presentation attack detection}. In \bibinfo{booktitle}{\emph{the IEEE/CVF Conference on Computer Vision and Pattern Recognition}}. \bibinfo{pages}{10023--10031}.
\newblock


\bibitem[Shen et~al\mbox{.}(2021)]%
        {shen2021agnostic}
\bibfield{author}{\bibinfo{person}{Zebang Shen}, \bibinfo{person}{Juan Cervino}, \bibinfo{person}{Hamed Hassani}, {and} \bibinfo{person}{Alejandro Ribeiro}.} \bibinfo{year}{2021}\natexlab{}.
\newblock \showarticletitle{An agnostic approach to federated learning with class imbalance}. In \bibinfo{booktitle}{\emph{International Conference on Learning Representations}}.
\newblock


\bibitem[Shi et~al\mbox{.}(2021)]%
        {shi2021gradient}
\bibfield{author}{\bibinfo{person}{Yuge Shi}, \bibinfo{person}{Jeffrey Seely}, \bibinfo{person}{Philip~HS Torr}, \bibinfo{person}{N Siddharth}, \bibinfo{person}{Awni Hannun}, \bibinfo{person}{Nicolas Usunier}, {and} \bibinfo{person}{Gabriel Synnaeve}.} \bibinfo{year}{2021}\natexlab{}.
\newblock \showarticletitle{Gradient matching for domain generalization}.
\newblock \bibinfo{journal}{\emph{arXiv preprint arXiv:2104.09937}} (\bibinfo{year}{2021}).
\newblock


\bibitem[Sun et~al\mbox{.}(2023)]%
        {pmlr-v189-sun23a}
\bibfield{author}{\bibinfo{person}{Yuwei Sun}, \bibinfo{person}{Ng Chong}, {and} \bibinfo{person}{Hideya Ochiai}.} \bibinfo{year}{2023}\natexlab{}.
\newblock \showarticletitle{Feature Distribution Matching for Federated Domain Generalization}. In \bibinfo{booktitle}{\emph{Asian Conference on Machine Learning}}, Vol.~\bibinfo{volume}{189}. \bibinfo{publisher}{PMLR}, \bibinfo{pages}{942--957}.
\newblock


\bibitem[Uddin et~al\mbox{.}(2023)]%
        {uddin2023arfl}
\bibfield{author}{\bibinfo{person}{Md~Palash Uddin}, \bibinfo{person}{Yong Xiang}, \bibinfo{person}{Borui Cai}, \bibinfo{person}{Xuequan Lu}, \bibinfo{person}{John Yearwood}, {and} \bibinfo{person}{Longxiang Gao}.} \bibinfo{year}{2023}\natexlab{}.
\newblock \showarticletitle{ARFL: Adaptive and Robust Federated Learning}.
\newblock \bibinfo{journal}{\emph{IEEE Transactions on Mobile Computing}} (\bibinfo{year}{2023}).
\newblock


\bibitem[Uddin et~al\mbox{.}(2024)]%
        {10234583}
\bibfield{author}{\bibinfo{person}{Md~Palash Uddin}, \bibinfo{person}{Yong Xiang}, \bibinfo{person}{Borui Cai}, \bibinfo{person}{Xuequan Lu}, \bibinfo{person}{John Yearwood}, {and} \bibinfo{person}{Longxiang Gao}.} \bibinfo{year}{2024}\natexlab{}.
\newblock \showarticletitle{ARFL: Adaptive and Robust Federated Learning}.
\newblock \bibinfo{journal}{\emph{IEEE Transactions on Mobile Computing}} \bibinfo{volume}{23}, \bibinfo{number}{5} (\bibinfo{year}{2024}), \bibinfo{pages}{5401--5417}.
\newblock
\urldef\tempurl%
\url{https://doi.org/10.1109/TMC.2023.3310248}
\showDOI{\tempurl}


\bibitem[Wang et~al\mbox{.}(2021)]%
        {wang2021proactive}
\bibfield{author}{\bibinfo{person}{Dong Wang}, \bibinfo{person}{Yuewei Yang}, \bibinfo{person}{Chenyang Tao}, \bibinfo{person}{Zhe Gan}, \bibinfo{person}{Liqun Chen}, \bibinfo{person}{Fanjie Kong}, \bibinfo{person}{Ricardo Henao}, {and} \bibinfo{person}{Lawrence Carin}.} \bibinfo{year}{2021}\natexlab{}.
\newblock \bibinfo{title}{Proactive Pseudo-Intervention: Causally Informed Contrastive Learning For Interpretable Vision Models}.
\newblock
\newblock
\showeprint[arxiv]{2012.03369}~[cs.CV]


\bibitem[Wang et~al\mbox{.}(2022)]%
        {wang2022domain}
\bibfield{author}{\bibinfo{person}{Jingye Wang}, \bibinfo{person}{Ruoyi Du}, \bibinfo{person}{Dongliang Chang}, {and} \bibinfo{person}{Zhanyu Ma}.} \bibinfo{year}{2022}\natexlab{}.
\newblock \showarticletitle{Domain generalization via frequency-based feature disentanglement and interaction}.
\newblock \bibinfo{journal}{\emph{CoRR}} (\bibinfo{year}{2022}).
\newblock


\bibitem[Wang et~al\mbox{.}(2020a)]%
        {NEURIPS2020_564127c0}
\bibfield{author}{\bibinfo{person}{Jianyu Wang}, \bibinfo{person}{Qinghua Liu}, \bibinfo{person}{Hao Liang}, \bibinfo{person}{Gauri Joshi}, {and} \bibinfo{person}{H.~Vincent Poor}.} \bibinfo{year}{2020}\natexlab{a}.
\newblock \showarticletitle{Tackling the Objective Inconsistency Problem in Heterogeneous Federated Optimization}. In \bibinfo{booktitle}{\emph{Advances in Neural Information Processing Systems}}, Vol.~\bibinfo{volume}{33}. \bibinfo{pages}{7611--7623}.
\newblock


\bibitem[Wang et~al\mbox{.}(2020b)]%
        {wang2020learning}
\bibfield{author}{\bibinfo{person}{Shujun Wang}, \bibinfo{person}{Lequan Yu}, \bibinfo{person}{Caizi Li}, \bibinfo{person}{Chi-Wing Fu}, {and} \bibinfo{person}{Pheng-Ann Heng}.} \bibinfo{year}{2020}\natexlab{b}.
\newblock \showarticletitle{Learning from extrinsic and intrinsic supervisions for domain generalization}. In \bibinfo{booktitle}{\emph{the European Conference on Computer Vision}}. Springer, \bibinfo{pages}{159--176}.
\newblock


\bibitem[Xu et~al\mbox{.}(2022)]%
        {xu2022closing}
\bibfield{author}{\bibinfo{person}{An Xu}, \bibinfo{person}{Wenqi Li}, \bibinfo{person}{Pengfei Guo}, \bibinfo{person}{Dong Yang}, \bibinfo{person}{Holger~R Roth}, \bibinfo{person}{Ali Hatamizadeh}, \bibinfo{person}{Can Zhao}, \bibinfo{person}{Daguang Xu}, \bibinfo{person}{Heng Huang}, {and} \bibinfo{person}{Ziyue Xu}.} \bibinfo{year}{2022}\natexlab{}.
\newblock \showarticletitle{Closing the generalization gap of cross-silo federated medical image segmentation}. In \bibinfo{booktitle}{\emph{the IEEE/CVF Conference on Computer Vision and Pattern Recognition}}. \bibinfo{pages}{20866--20875}.
\newblock


\bibitem[Xu et~al\mbox{.}(2023)]%
        {xu2023multi}
\bibfield{author}{\bibinfo{person}{Mingjun Xu}, \bibinfo{person}{Lingyun Qin}, \bibinfo{person}{Weijie Chen}, \bibinfo{person}{Shiliang Pu}, {and} \bibinfo{person}{Lei Zhang}.} \bibinfo{year}{2023}\natexlab{}.
\newblock \showarticletitle{Multi-view adversarial discriminator: Mine the non-causal factors for object detection in unseen domains}. In \bibinfo{booktitle}{\emph{the IEEE/CVF Conference on Computer Vision and Pattern Recognition}}. \bibinfo{pages}{8103--8112}.
\newblock


\bibitem[Xu et~al\mbox{.}(2021)]%
        {xu2021fourier}
\bibfield{author}{\bibinfo{person}{Qinwei Xu}, \bibinfo{person}{Ruipeng Zhang}, \bibinfo{person}{Ya Zhang}, \bibinfo{person}{Yanfeng Wang}, {and} \bibinfo{person}{Qi Tian}.} \bibinfo{year}{2021}\natexlab{}.
\newblock \showarticletitle{A fourier-based framework for domain generalization}. In \bibinfo{booktitle}{\emph{the IEEE/CVF Conference on Computer Vision and Pattern Recognition}}. \bibinfo{pages}{14383--14392}.
\newblock


\bibitem[Ye et~al\mbox{.}(2023)]%
        {ye2023fedfm}
\bibfield{author}{\bibinfo{person}{Rui Ye}, \bibinfo{person}{Zhenyang Ni}, \bibinfo{person}{Chenxin Xu}, \bibinfo{person}{Jianyu Wang}, \bibinfo{person}{Siheng Chen}, {and} \bibinfo{person}{Yonina~C Eldar}.} \bibinfo{year}{2023}\natexlab{}.
\newblock \showarticletitle{Fed{FM}: Anchor-based feature matching for data heterogeneity in federated learning}.
\newblock \bibinfo{journal}{\emph{IEEE Transactions on Signal Processing}} (\bibinfo{year}{2023}).
\newblock


\bibitem[Yoon et~al\mbox{.}(2021)]%
        {yoon2021federated}
\bibfield{author}{\bibinfo{person}{Jaehong Yoon}, \bibinfo{person}{Wonyong Jeong}, \bibinfo{person}{Giwoong Lee}, \bibinfo{person}{Eunho Yang}, {and} \bibinfo{person}{Sung~Ju Hwang}.} \bibinfo{year}{2021}\natexlab{}.
\newblock \showarticletitle{Federated continual learning with weighted inter-client transfer}. In \bibinfo{booktitle}{\emph{International Conference on Machine Learning}}. PMLR, \bibinfo{pages}{12073--12086}.
\newblock


\bibitem[Yu et~al\mbox{.}(2020)]%
        {yu2020bdd100k}
\bibfield{author}{\bibinfo{person}{Fisher Yu}, \bibinfo{person}{Haofeng Chen}, \bibinfo{person}{Xin Wang}, \bibinfo{person}{Wenqi Xian}, \bibinfo{person}{Yingying Chen}, \bibinfo{person}{Fangchen Liu}, \bibinfo{person}{Vashisht Madhavan}, {and} \bibinfo{person}{Trevor Darrell}.} \bibinfo{year}{2020}\natexlab{}.
\newblock \showarticletitle{Bdd100k: A diverse driving dataset for heterogeneous multitask learning}. In \bibinfo{booktitle}{\emph{the IEEE/CVF International Conference on Computer Vision}}. \bibinfo{pages}{2636--2645}.
\newblock


\bibitem[Yuan et~al\mbox{.}(2021)]%
        {yuan2021we}
\bibfield{author}{\bibinfo{person}{Honglin Yuan}, \bibinfo{person}{Warren Morningstar}, \bibinfo{person}{Lin Ning}, {and} \bibinfo{person}{Karan Singhal}.} \bibinfo{year}{2021}\natexlab{}.
\newblock \showarticletitle{What do we mean by generalization in federated learning?}
\newblock \bibinfo{journal}{\emph{arXiv preprint arXiv:2110.14216}} (\bibinfo{year}{2021}).
\newblock


\bibitem[Zhang et~al\mbox{.}(2023a)]%
        {10011632}
\bibfield{author}{\bibinfo{person}{Liling Zhang}, \bibinfo{person}{Xinyu Lei}, \bibinfo{person}{Yichun Shi}, \bibinfo{person}{Hongyu Huang}, {and} \bibinfo{person}{Chao Chen}.} \bibinfo{year}{2023}\natexlab{a}.
\newblock \showarticletitle{Federated Learning for IoT Devices With Domain Generalization}.
\newblock \bibinfo{journal}{\emph{IEEE Internet of Things Journal}} \bibinfo{volume}{10}, \bibinfo{number}{11} (\bibinfo{year}{2023}), \bibinfo{pages}{9622--9633}.
\newblock
\urldef\tempurl%
\url{https://doi.org/10.1109/JIOT.2023.3234977}
\showDOI{\tempurl}


\bibitem[Zhang et~al\mbox{.}(2021)]%
        {9710573}
\bibfield{author}{\bibinfo{person}{Lin Zhang}, \bibinfo{person}{Yong Luo}, \bibinfo{person}{Yan Bai}, \bibinfo{person}{Bo Du}, {and} \bibinfo{person}{Ling-Yu Duan}.} \bibinfo{year}{2021}\natexlab{}.
\newblock \showarticletitle{Federated Learning for Non-IID Data via Unified Feature Learning and Optimization Objective Alignment}. In \bibinfo{booktitle}{\emph{the IEEE/CVF International Conference on Computer Vision}}. \bibinfo{pages}{4400--4408}.
\newblock


\bibitem[Zhang et~al\mbox{.}(2022)]%
        {zhang2022semi}
\bibfield{author}{\bibinfo{person}{Ruipeng Zhang}, \bibinfo{person}{Qinwei Xu}, \bibinfo{person}{Chaoqin Huang}, \bibinfo{person}{Ya Zhang}, {and} \bibinfo{person}{Yanfeng Wang}.} \bibinfo{year}{2022}\natexlab{}.
\newblock \showarticletitle{Semi-supervised domain generalization for medical image analysis}. In \bibinfo{booktitle}{\emph{International Symposium on Biomedical Imaging}}. IEEE, \bibinfo{pages}{1--5}.
\newblock


\bibitem[Zhang et~al\mbox{.}(2023b)]%
        {zhang2023federated}
\bibfield{author}{\bibinfo{person}{Ruipeng Zhang}, \bibinfo{person}{Qinwei Xu}, \bibinfo{person}{Jiangchao Yao}, \bibinfo{person}{Ya Zhang}, \bibinfo{person}{Qi Tian}, {and} \bibinfo{person}{Yanfeng Wang}.} \bibinfo{year}{2023}\natexlab{b}.
\newblock \showarticletitle{Federated domain generalization with generalization adjustment}. In \bibinfo{booktitle}{\emph{the IEEE/CVF Conference on Computer Vision and Pattern Recognition}}. \bibinfo{pages}{3954--3963}.
\newblock


\end{thebibliography}

\end{document}